%% file: main.tex
\pdfoutput=1

\documentclass[11pt]{article}

\usepackage[]{latex/acl}

\usepackage{times}
\usepackage{latexsym}
\usepackage{amssymb} 
\usepackage{pifont} 
\include{commands}
\include{comments}

\usepackage[T1]{fontenc}

\usepackage[utf8]{inputenc}

\usepackage{microtype}
\usepackage{comment}

\usepackage{inconsolata}
\usepackage{marvosym}

\usepackage{multirow}

\title{AfriMTE and AfriCOMET: Enhancing COMET to Embrace Under-resourced African Languages}

\author{\normalsize 
Jiayi Wang$^{1}$, 
David Ifeoluwa Adelani$^{1,2}$, 
Sweta Agrawal$^{3,}$\thanks{~~The authors contribute equally to this work and are considered co-third authors.} , 
Marek Masiak$^{1,\ast,}$\thanks{~~Currently at the University of Oxford, UK.} ,
Ricardo Rei$^{4,5,6}$,\\
\textbf{\normalsize 
Eleftheria Briakou$^{3}$,
Marine Carpuat$^{3}$, 
Xuanli He$^{1}$,  
Sofia Bourhim$^{7}$,
Andiswa Bukula$^8$,}\\
\textbf{\normalsize 
Muhidin Mohamed$^{9}$, 
Temitayo Olatoye$^{10}$,
Tosin Adewumi$^{11}$,
Hamam Mokayed$^{11}$,} \\
\textbf{\normalsize
Christine Mwase$^{12}$,
Wangui Kimotho$^{2}$, 
Foutse Yuehgoh$^{13}$, 
Anuoluwapo Aremu$^{2}$,} \\
\textbf{\normalsize
Jessica Ojo$^{14,2}$,
Shamsuddeen Hassan Muhammad$^{15,2,29}$, 
Salomey Osei$^{16,2}$,} \\
\textbf{\normalsize
Abdul-Hakeem Omotayo$^{17,2}$,
Chiamaka Chukwuneke$^{18,2}$, 
Perez Ogayo$^{2}$, 
Oumaima Hourrane$^{2}$,} \\
\textbf{\normalsize
Salma El Anigri$^{19}$,
Lolwethu Ndolela$^{2}$,  
Thabiso Mangwana$^{2}$,  
Shafie Abdi Mohamed$^{20}$,} \\
\textbf{\normalsize
Ayinde Hassan$^{21}$,
Oluwabusayo Olufunke Awoyomi$^{22}$, 
Lama Alkhaled$^{11}$, 
Sana Al-Azzawi$^{11}$,} \\
\textbf{\normalsize
Naome A. Etori$^{23}$,
Millicent Ochieng$^{24}$,  
Clemencia Siro$^{25}$, 
Samuel Njoroge$^{26}$, 
Eric Muchiri$^{2}$,} \\
\textbf{\normalsize
Wangari Kimotho$^{27}$,
Lyse Naomi Wamba Momo$^{28}$, 
Daud Abolade$^{2}$,  
Simbiat Ajao$^{2}$,} \\ 
\textbf{\normalsize
Iyanuoluwa Shode$^{2}$,  
Ricky Macharm$^{2}$, 
Ruqayya Nasir Iro$^{29}$,
Saheed S. Abdullahi$^{30,31}$,}\\ 
\textbf{\normalsize
Stephen E. Moore$^{32,33}$, 
Bernard Opoku$^{34,2}$, 
Zainab Akinjobi$^{35,2}$,
Abeeb Afolabi$^{2}$,}\\ 
\textbf{\normalsize
Nnaemeka Obiefuna$^{2}$, 
Onyekachi Raphael Ogbu$^{2}$, 
Sam Brian$^{2}$,
Verrah Akinyi Otiende$^{36}$,}\\ 
\textbf{\normalsize
Chinedu Emmanuel Mbonu$^{37}$, 
Sakayo Toadoum Sari$^{38}$, 
Yao Lu$^{1}$, 
Pontus Stenetorp$^{1}$} \\ 
\footnotesize
$^1$University College London, UK, 
$^2$Masakhane NLP,
$^3$University of Maryland, USA, \\
\footnotesize
$^4$Unbabel,
$^5$Instituto Superior Técnico, 
$^6$INESC-ID, 
$^7$ENSIAS, Morocco, 
$^8$SADiLaR, South Africa, \\
\footnotesize
$^9$Aston University, UK, 
$^{10}$University of Eastern Finland, Finland, 
$^{11}$Luleå University of Technology, Sweden, \\
\footnotesize
$^{12}$Fudan University, China, 
$^{13}$Conservatoire National des Arts et Métiers, France, 
$^{14}$Lelapa AI, South Africa, \\
\footnotesize
$^{15}$Imperial College London, UK, 
$^{16}$University of Deusto, Spain,
$^{17}$University of California, USA, 
$^{18}$Lancaster University, UK,\\
\footnotesize
$^{19}$Mohammed V University, Morocco,
$^{20}$Jamhuriya University Of Science and Technology, Somalia, \\
\footnotesize
$^{21}$LAUTECH, Nigeria,
$^{22}$The College of Saint Rose, USA, 
$^{23}$University of Minnesota -Twin Cities, USA, \\
\footnotesize
$^{24}$Microsoft Africa Research Institute, 
$^{25}$University of Amsterdam, Netherlands,
$^{26}$The Technical University of Kenya, \\
\footnotesize
$^{27}$AIMS, Cameroon, 
$^{28}$KU Leuven, Belgium, 
$^{29}$HausaNLP, 
$^{30}$SIAT-CAS, China, 
$^{31}$Kaduna State University, Nigeria, \\
\footnotesize 
$^{32}$University of Cape Coast, Ghana, 
$^{33}$Ghana NLP, 
$^{34}$Kwame Nkrumah University of Science and Technology, Ghana, \\ 
\footnotesize 
$^{35}$New Mexico State University, USA, 
$^{36}$USIU-Africa, 
$^{37}$UNIZIK, Nigeria, 
$^{38}$AIMS, Senegal \\
\texttt{Corresponding emails: \{ucabj45,d.adelani\}@ucl.ac.uk,} \\ 
\texttt{p.stenetorp@cs.ucl.ac.uk} \\ 
}

\begin{document}

\maketitle
\begin{abstract}
Despite the recent progress on scaling multilingual machine translation (MT) to several under-resourced African languages, accurately measuring this progress remains challenging, since evaluation is often performed on $n$-gram matching metrics such as BLEU, which typically show a weaker correlation with human judgments.
Learned metrics such as COMET have higher correlation; however, the lack of evaluation data with human ratings for under-resourced languages, complexity of annotation guidelines like Multidimensional Quality Metrics~(MQM), and limited language coverage of multilingual encoders have hampered their applicability to African languages.
In this paper, we address these challenges by creating high-quality human evaluation data with simplified MQM guidelines for error detection and direct assessment~(DA) scoring for 13 typologically diverse African languages.
Furthermore, we develop \textsc{AfriCOMET}: COMET evaluation metrics for African languages by leveraging DA data from well-resourced languages and an African-centric multilingual encoder~(AfroXLM-R) to create the state-of-the-art MT evaluation metrics for African languages with respect to Spearman-rank correlation with human judgments~($0.441$).
\end{abstract}

\input{introduction}
\input{dataset}

\input{african_comet}

\input{ref_free_model}

\section{Additional Evaluation}
Additional evaluations have been conducted on the generalization of our AfriCOMET and AfriCOMET-QE systems to other datasets. Please refer to Appendix~\ref{subsec:appendix b},~\ref{subsec:appendix c} and~\ref{subsec:generalization} for details.

\section{Conclusion}
This study tackles the challenges of enhancing the COMET metric for various under-resourced African languages. We simplify the MQM annotation guidelines for non-expert evaluators, create an MT evaluation dataset, \textsc{AfriMTE}, covering 13 typologically diverse African languages, and establish benchmark MT evaluation~(\textsc{AfriCOMET}) and reference-free QE~(\textsc{AfriCOMET-QE}) systems. Our findings show the feasibility of employing transfer learning from well-resourced DA data and an African-centric multilingual pre-trained encoder, AfroXLM-R, for building MT evaluation and QE models for African languages. 

\input{paper_limitation}

\section*{Acknowledgments}
David Adelani acknowledges the support of DeepMind Academic Fellowship programme. 
Ricardo Rei is supported by the European Union's Horizon Europe Research and Innovation Actions (UTTER: contract 101070631) and by the Portuguese Recovery and Resilience Plan through project C645008882-00000055 (Center for Responsible AI). 
Pontus Stenetorp would like to acknowledge the helpful proofing feedback from several viewers while finalizing the submission.
We are grateful to Prof. Antonios Anastasopoulos from George Mason University for releasing the ``WMT African'' dataset for our experiments. 
This work is supported in part by Oracle Cloud credits and related resources provided by Oracle.
This work is also supported in part by Microsoft Research via their Accelerate Foundation Models Research Grant. 
Finally, we are grateful to OpenAI for providing Masakhane with API credits through their Researcher Access API program for the evaluation of the GPT-4 large language models.




\bibliography{anthology,custom}

\appendix
\input{sec-app}

\end{document}

%% file: commands.tex
\usepackage[inline]{enumitem}
\usepackage[utf8]{inputenc} 
\usepackage[T1]{fontenc}    
\usepackage[english]{babel}
\usepackage{hyperref}       
\usepackage{url}            
\usepackage{amsfonts}       
\usepackage{nicefrac}       
\usepackage{microtype}      
\usepackage{booktabs}
\usepackage{amsfonts}
\usepackage{amsmath}
\usepackage{multirow}
\usepackage{graphicx}
\usepackage{enumitem}
\usepackage{subcaption}
\usepackage{caption}
\usepackage{rotating}
\usepackage[normalem]{ulem}
\usepackage{amssymb}
\usepackage{colortbl}
\usepackage{linguex}

\usepackage{siunitx} 
\usepackage{multirow, makecell}
\usepackage{pifont}
\usepackage{siunitx} 
\usepackage{tabularx}
\usepackage[all]{nowidow}
\usepackage{colortbl}
\usepackage{nicematrix,tikz}
\usepackage{xcolor}

\definecolor{hotpink}{HTML}{EF7C8E}
\definecolor{tiffanyblue}{HTML}{A0E7E5}
\definecolor{mint}{HTML}{B4F8C8}
\definecolor{paleyellow}{HTML}{FBE7C6}
\definecolor{rosewater}{HTML}{D8A7B1}
\definecolor{cream}{HTML}{FAE8E0}

%% file: comments.tex
\usepackage{xcolor}



%% file: introduction.tex
\section{Introduction}
Recent advances in machine translation~(MT) have focused on scaling multilingual translation models and evaluation data to hundreds of languages, including multiple under-resourced languages~\citep{m2m100_fan,team2022NoLL,Bapna2022BuildingMT,Kudugunta2023MADLAD400AM}.
However, measuring the progress made for these under-resourced languages accurately is difficult, since popular $n$-gram matching metrics, such as BLEU~\citep{papineni-etal-2002-bleu}, METEOR~\citep{banerjee-lavie-2005-meteor}, and ChrF~\citep{popovic-2015-chrf}, fail to capture semantic similarity beyond the lexical level~\citep{Zhang_020BERTScore, rei-etal-2020-comet, sai-b-etal-2023-indicmt}.
Variants of these metrics have been developed when scaling to various languages such as spBLEU~\citep{m2m100_fan}, but they often achieve worse correlation to human judgements~\citep{freitag-etal-2022-results} when compared to embedding-based metrics like BERTScore~\citep{Zhang_020BERTScore}, and learned metrics such as 
COMET~\citep{rei-etal-2020-comet}.

While embedding-based metrics are currently favored for evaluation in MT~\citep{freitag-etal-2022-results}, the application of these metrics to under-resourced languages faces three challenges: (1) lack of high-quality training and evaluation data 
significantly hampers the development of reliable metrics; (2) the complexity of the Multidimensional Quality Metrics (MQM) framework~\citep{Lommel2014MultidimensionalQM} presents a steep learning curve for non-expert bilingual evaluators, complicating the process of obtaining accurate human assessments; 
and (3) the limited language coverage of multilingual large language models such as XLM-R~\citep{conneau-etal-2020-unsupervised} restricts their applicability to various low-resource languages~\citep{alabi-etal-2022-adapting}.
%

To address these challenges, recent work have utilized the Direct Assessment~(DA) scoring annotations~\citep{graham-etal-2013-continuous} collected by the organizers of WMT~\citep{rei2022comet} and leveraged the transfer learning capabilities of multilingual encoders to evaluate unseen languages~\citep{rei-etal-2022-cometkiwi,zerva-etal-2022-findings}. However, the dearth of evaluation data for under-resourced languages such as African languages still remains a significant hurdle in validating these methods. What is worse, as~\citet{rei-etal-2020-comet} highlighted, the performance of these approaches is often unpredictable for languages that were not included in the pre-training phase of multilingual language models.

In this paper, we address these challenges by enhancing the state-of-the-art COMET evaluation metric~\citep{rei2022comet} to various under-resourced African languages.
To overcome the scarcity of evaluation datasets, we create \textsc{AfriMTE}---a human evaluation dataset focusing on MT adequacy and fluency evaluation for 13 typologically diverse African languages.
This is achieved through a participatory research methodology, ensuring a comprehensive and representative data collection process~\citep{nekoto-etal-2020-participatory}.
In addressing the complexities inherent in the MQM framework, 
we develop a simplified version that aligns with the tenets of Direct Assessment~(DA) and is tailored specifically for non-expert evaluators, aiming to augment both usability and accessibility, thereby rendering the evaluation process more accessible to a wider spectrum of evaluators.

Finally, we develop the first COMET model designed for MT evaluation for African languages. 
Additionally, we introduce the first translation quality estimation~(QE) model for African languages, which operates translation quality estimation without requiring reference translations, setting a new benchmark in the QE field~\citep{fan2019bilingual, specia-etal-2020-findings-wmt, specia2021findings, wang2021qemind}.

To summarize, our contributions are as follows:
(1)
we propose simplified MQM evaluation guidelines tailored for non-expert translators;
(2)
to support the application of our guidelines, we develop a specialized annotation tool;
(3)
we develop a high-quality human evaluation dataset focusing on machine translation adequacy and fluency for 13 typologically diverse African languages;
(4)
we establish benchmark systems for MT Evaluation and Quality Estimation by employing transfer learning techniques from existing, well-resourced DA data and utilizing an African-centric multilingual pre-trained language model;
(5)
to foster ongoing research in the domain of African machine translation evaluation, we will release all evaluation datasets, code, and models publicly.\footnote{The resources will be publicly available at \url{https://github.com/masakhane-io/africomet}.}

%% file: dataset.tex
\section{\textsc{AfriMTE}: African Machine Translation Evaluation Dataset}
This section details the data and machine translation engines used for annotation, outlines our annotation guidelines and procedure, describes the data quality assurance process, and presents a quantitative analysis of the collected data.

\subsection{Dataset and MT Engine}
Our annotation work concentrates on the~\textbf{dev} and~\textbf{devtest} subsets from the FLORES-200 dataset~\citep{team2022NoLL}. This is a multi-way parallel dataset designed to enhance MT for low-resource languages. Flores-200 source texts were sampled from English Wikipedia articles and 
reference translations into target languages were produced by 
professional translators. We focus on 13 languages pairs (LPs): Darija-French (ary-fra), English-Egyptian Arabic (eng-arz), English-French (eng-fra)---a control LP, 
English-Hausa (eng-hau), English-Igbo (eng-ibo), English-Kikuyu (eng-kik), English-Luo (eng-luo), English-Somali (eng-som), English-Swahili (eng-swh), English-Twi (eng-twi), English-isiXhosa (eng-xho), English-Yoruba (eng-yor), and Yoruba-English (yor-eng). 
%
Moreover, 
we extend our annotation collection to include domain-specific texts from News, TED talks, Movies, and IT domains for English-Yoruba translations, which were established in prior research by~\citet{adelani2021effect} and~\citet{shode2022yosm}, ensuring a comprehensive and domain-varied evaluation.
We provide the information of language family groups that our targeted African languages belong to in Table~\ref{tab:language_family} of Appendix~\ref{sec:appendix_a}.

To acquire MT outputs, we employ two open-source MT engines: NLLB-200 (600M)~\citep{team2022NoLL} and M2M-100 (418M)~\citep{fan2021beyond}. For eng-fra and eng-swh, we generate translations using M2M-100, while for all other LPs, we utilize NLLB-200. 
This decision is based on the exceptional proficiency of NLLB-200 translations for eng-fra and eng-swh, where our evaluators found them to be almost error-free during the example annotation training phase. While for some LPs such as eng-xho and eng-yor TED talks, despite their overall high translation quality at the sentence level, our evaluators noted minor error at the word level, as shown in Figures~\ref{fig:span_plot} and~\ref{fig:span_plot_flu} of Appendix~\ref{sec:appendix_a}. Therefore, we retain the NLLB-200 engine for these languages. The presence of such slight errors provides an opportunity to assess the robustness and sensitivity of our developed metrics in situations with minimal translation errors. When generating translations, we consistently use a beam size of 5 for both engines. 

In the FLORES-200 dataset, we sample 270 and 250 sentences respectively from the dev and devtest sets. The sampling reflects the averaged SacreBLEU~\citep{post-2018-call} scores for both high-quality and lower-quality translations across 21 language pairs, ensuring a balanced representation of translation effectiveness.\footnote{Note that our project initially included more LPs, but due to limited evaluators, 13 remained in AfriMTE.} Finally, our annotation datasets are structured as triple parallel, comprising <source, machine translation, reference> for all LPs.


\begin{figure*}[h]
\centering
\begin{subfigure}{0.90\linewidth}
  \centering
\includegraphics[width=0.90\textwidth]{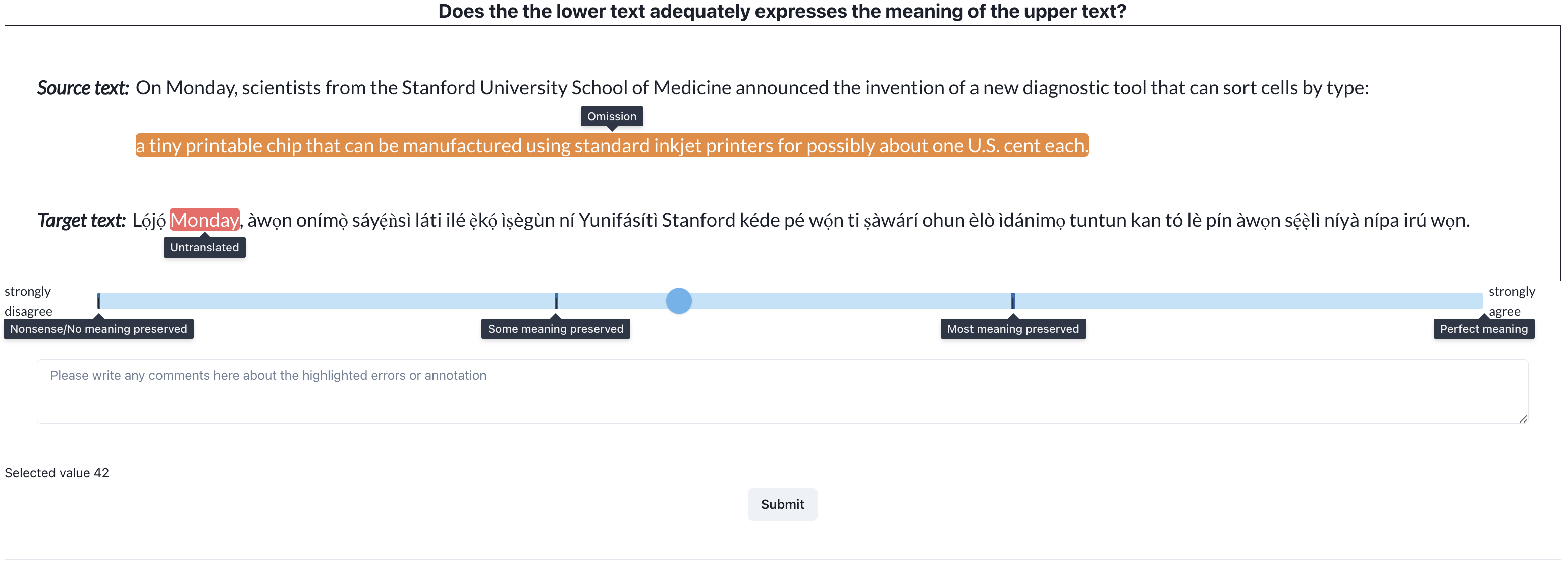}
\end{subfigure}%
\vspace{-3mm}
\caption{The screenshot of the user interface with an adequacy annotated task comprising the source sentence and its corresponding translation in English-Yoruba.} 
\label{fig:UI}
\end{figure*}

\begin{figure}[t]
\centering
\includegraphics[width=1.0\columnwidth]{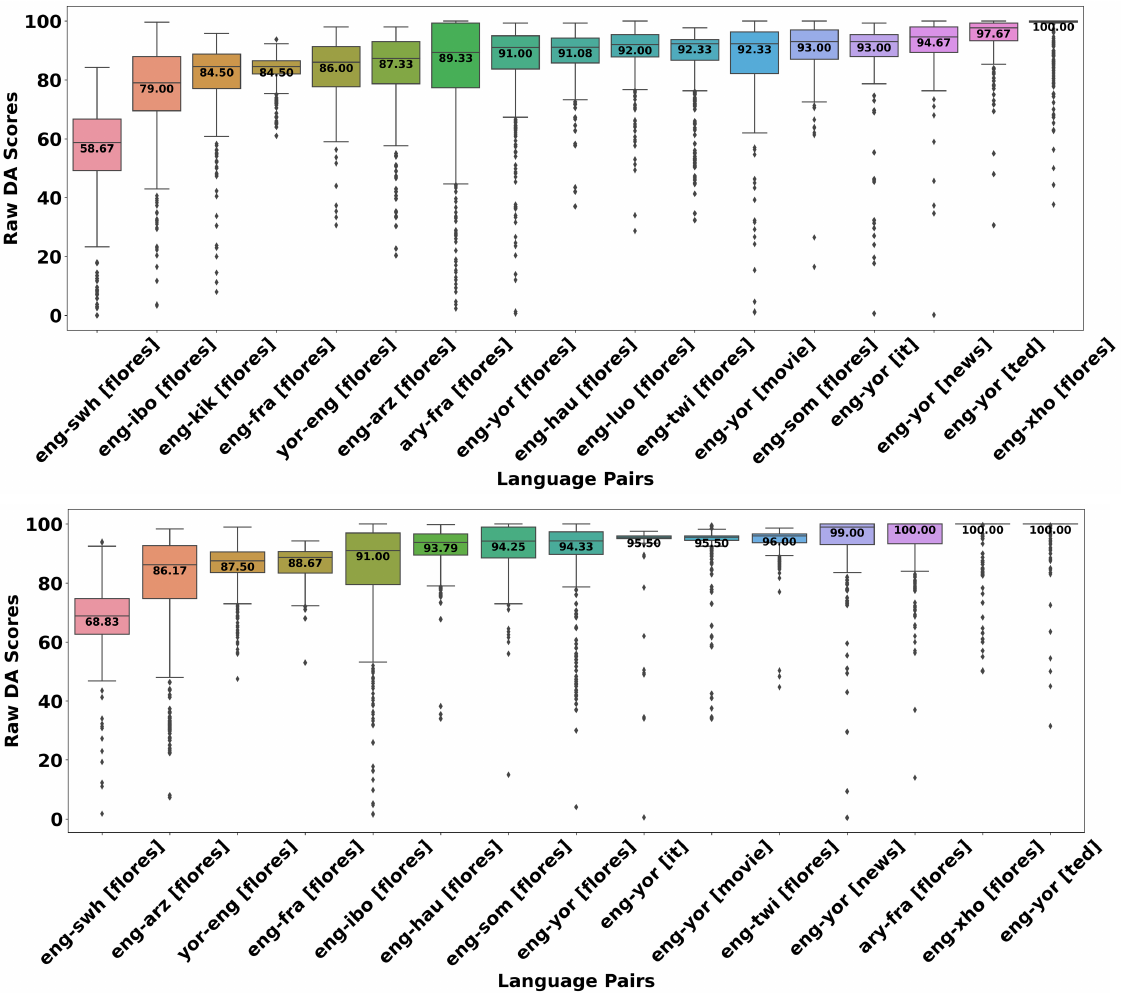}
\caption{Translation quality of~\textbf{all} qualified annotated translations as measured by raw DA scores across all language pairs and domains 
in ascending order, with medians displayed in the plot for~\textbf{adequacy}~(upper) and~\textbf{fluency}~(lower).
} 
\label{fig:quality_DA_raw_for_all}
\end{figure}

\subsection{Annotation Guidelines and Tool}
This section presents our annotation guidelines and introduces the annotation tool.



\subsubsection{Annotation Guidelines}
Recent findings~\citep{freitag2021experts} have indicated that crowd-sourced DA annotations tend to be inconsistent in assessing the quality of high-performing MT systems.
This led us to consider adopting the standardized MQM framework~\citep{Lommel2014MultidimensionalQM}---an extensive method for assessing translation quality by defining various error dimensions collected alongside error severity.
However, its complex nature presents a learning hurdle for non-expert evaluators, which was recognized during our annotation training phase.
Research by \citet{bentivogli2018machine} and \citet{chatzikoumi2020evaluate} shows that while DA has traditionally been used for both translation adequacy and fluency, it currently focuses more on adequacy.
Moreover, \citet{graham-etal-2013-continuous, graham2017can} suggests employing DA to evaluate both aspects on a 100-point scale.
Drawing upon these findings, we propose a simplified MQM guideline focusing on translation adequacy, combining translation accuracy error detection with DA scoring for ease of use by non-expert evaluators.
Similarly, we create a distinct MQM guideline for translation fluency, combining translation fluency error detection with DA scoring.

Our evaluators assess translation adequacy and fluency separately, both through a two-dimensional approach: error highlighting and overall DA scoring. In assessing adequacy, evaluators review both the source and translated texts, highlighting errors categorized as ``Addition'', ``Omission'', ``Mistranslation'', and ``Untranslated''. During the fluency assessment, evaluators focus solely on the translated text, pinpointing errors in ``Grammar'', ``Spelling'', ``Typography'', and ``Unintelligible''. 
The specific error definitions are adapted from the original MQM framework.\footnote{\url{https://themqm.org}}

Upon completing error highlightings, evaluators use the DA guidelines to assign a score between 0 and 100, reflecting the overall quality of adequacy or fluency. We are motivated by the DA+SQM framework~\citep{kocmi2023findings} for our DA guidelines, where we additionally bucket scores based on specific levels to reduce subjectivity. Specifically, in these scales, ``0'' is defined as a ``Nonsense/No meaning preserved'' translation for adequacy or an ``Incomprehensible'' translation for fluency, while ``100'' signifies ``Perfect meaning'' for adequacy or ``Fluent and natural'' for fluency. 
In addition, there are two intermediate score levels within either rating scale: one at ``34'' and the another at ``67''. Details of the adequacy and fluency guidelines are illustrated in Figure~\ref{fig:ade_guideline} and~\ref{fig:flu_guideline} of Appendix~\ref{sec:appendix_a}, either with two sections: the error highlighting guidelines and the DA scoring guidelines.

\subsubsection{Annotation Tool}
To collect annotations following our tailored annotation guidelines, 
we extend an internal annotation tool to suit our needs.\footnote{\url{https://github.com/marek357/annotation-tool-frontend}}
Various features have been added, including presenting evaluators with annotation guidelines, adapting the interface to accommodate the error span highlighting and DA scoring functions, and exporting annotations appropriately. The customized tool provides a user-friendly interface designed for machine translation evaluation tasks. 
%
A screenshot of the annotation interface is displayed in Figure~\ref{fig:UI}, where every evaluator can work independently.


\subsection{Annotation Quality Assurance}
\label{sec:quality}
We implement a stringent evaluation protocol for each translation, involving a minimum of \textbf{two} bilingual native speakers as evaluators, each with a Bachelor's degree or higher.
They are encouraged to highlight specific error spans first and then provide a relevant DA score before submission. In preparation, each evaluator annotates 20 examples, and we organize a discussion among the evaluators to review annotations and address any assessment inconsistencies. This preliminary step is designed to familiarize evaluators with the guidelines and the dataset contexts. 
Some annotators assign low DA scores but lack any corresponding error span highlighting. Hence, in the following data analysis of the correlation between error counts and overall DA scoring, we will exclude such annotations.

Upon completing annotations, we gather data and exclude any with DA score discrepancies exceeding 34 points, as per our guidelines. This threshold
is critical for ensuring the reliability of our annotations.
To reduce bias among evaluators, we normalize DA scores at the evaluator level to get z-scores, and then average z-scores across evaluators to obtain the final score of each translation.
We present the counts of qualified translation annotations within the dev and devtest sets in Table~\ref{tab:overall_count} and~\ref{tab:overall_count_fluency} in Appendix~\ref{sec:appendix_a}.
%

To further validate annotation consistency, we apply the inter-annotator agreement~(IAA) method~\citep{pavlick-tetreault-2016-empirical}. Each annotation instance is randomly split, with one as Annotator 1 and the average of others as Annotator 2. We compute the Pearson correlation 
between these two groups, repeating this process 100 times. The average IAA scores are 0.797 for adequacy and 0.748 for fluency, demonstrating strong consistency among evaluators.

\subsection{Quantitative Analysis of Annotations}
\paragraph{Overall Translation Quality} 
We show the distributions of raw DA scores across all LPs in Figure~\ref{fig:quality_DA_raw_for_all}.\footnote{We still add domain-specific eng-yor annotations in the plot.} Notably, eng-swh translations generated by M2M-100 exhibit the lowest translation adequacy and fluency~(median DA: 58.67 and 68.83), whereas eng-xho translations by NLLB-200 score the highest~(median DA: 100 for both). Moreover, ary-fra translations have the highest variance in adequacy, while eng-arz and eng-ibo in fluency. The perceived low adequacy and fluency scores observed for the control LP, eng-fra, are consistent with the MT engine employed for it in our study. 



\paragraph{Error Counts vs. DA score} 
Equipped with annotations predominantly comprising both overall DA scores and detection of fine-grained error spans, we aim to investigate the correlation between these two aspects. As previously mentioned in Section~\ref{sec:quality}, some annotations have low DA scores without error span highlighting. Therefore, we exclude annotations with DA scores under 80 lacking error span highlighting. After this filtration, we present the counts of error words per category and their sentence-level DA scores in Figures~\ref{fig:span_plot} and~\ref{fig:span_plot_flu} respectively of Appendix~\ref{sec:appendix_a} for adequacy and fluency.

Mistranslation is the predominant error impacting adequacy, significantly contributing to lower DA scores. Interestingly, eng-yor Movie translations exhibit a higher incidence of Omission errors, whereas eng-yor IT translations are more prone to Addition errors. Unintelligible is the most common error for fluency except for eng-swh, eng-som, eng-hau. This is consistent in eng-yor domain-specific translations except for the Movie domain.


\input{tables/corr_table_jiayi}

In order to better understand how word-level errors influence judgments at the sentence level, we calculate and report Pearson, Spearman-rank, and Kendall-rank correlation coefficients between counts within each error category and corresponding scores~(raw DA scores and normalized z-scores) in Table~\ref{tab:corr_features_jiayi}.
These coefficients suggest that Mistranslation and Unintelligible, as the most prevalent error categories for adequacy and fluency, exhibit moderate to high negative correlations with raw DA scores, indicating their significant influence on the sentence-level DA judgements from evaluators.
Moreover, for both adequacy and fluency, the total and average error counts per reference translation length negatively correlate with the raw DA scores and the normalized z-scores, further affirming the significance of our simplified MQM guidelines.



%% file: tables/corr_table_jiayi.tex
\begin{table}[t]
\centering
\small
 \setlength\tabcolsep{4pt}
\scalebox{0.75}{
\begin{tabular}{l|cc|cc|cc}
 \toprule
 \textbf{\textsc{Criteria}}  & 
 \multicolumn{2}{c}{\textbf{\textsc{Pearson}}} &
 \multicolumn{2}{c}{\textbf{\textsc{Spearman}}} & 
 \multicolumn{2}{c}{\textbf{\textsc{Kendall}}} \\
 & DA score & Z-score & DA score & Z-score & DA score & Z-score \\
\midrule
Mistranslation	& -0.478 & -0.398 & -0.675 & -0.544 & -0.546 & -0.422 \\
Omission & -0.180 & -0.196 & -0.318 & -0.304 & -0.263 & -0.246 \\
Addition & -0.236 & -0.291 & -0.207 & -0.211 & -0.172 & -0.172 \\
Untranslated & -0.091 & -0.101 & -0.156 & -0.119 & -0.130 & -0.097 \\
\midrule
Total Error	& -0.467 & -0.479 & -0.791 & \textbf{-0.687} & \textbf{-0.640} & \textbf{-0.533} \\
Avg. Error & \textbf{-0.490} & \textbf{-0.490} & \textbf{-0.792} & -0.681 & -0.627 & -0.516 \\
\midrule
\midrule
Grammar	& -0.322 & -0.191 & -0.422 & -0.279 & -0.355 & -0.223 \\
Spelling & -0.042 & -0.075 & -0.078 & -0.103 & -0.066 & -0.084 \\
Typography & -0.158 & -0.257 & -0.180 & -0.193 & -0.153 & -0.157 \\
Unintelligible & -0.442 & -0.470 & -0.466 & -0.354 & -0.396 & -0.286 \\
\midrule
Total Error	& \textbf{-0.546} & \textbf{-0.577} & \textbf{-0.685} & \textbf{-0.536} & \textbf{-0.576} & \textbf{-0.421} \\
Avg. Error & -0.509 & -0.539 & \textbf{-0.685} & -0.527 & -0.568 & -0.409 \\
  \bottomrule
 \end{tabular}
}
\vspace{-3mm}
\caption{Correlations between error counts and sentence-level scores across error categories for~\textbf{adequacy} (upper) and~\textbf{fluency} (lower) respectively. ``Avg. Error'' refers to the average error counts per reference length.}
\label{tab:corr_features_jiayi}
\vspace{-0.5cm}
\end{table}

%% file: african_comet.tex

\section{\textsc{AfriCOMET}: Benchmark Systems}
\label{sec:afriCOMET}
In this section, we will describe the development of our MT evaluation systems for African languages using~\textsc{AfriMTE}. The primary objective of our modeling is to predict the normalized adequacy DA score. Our investigation centers around three key questions: (1) the feasibility of constructing an MT evaluation system that leverages transfer learning from other languages to African languages, (2) the impact of using African language-enhanced pre-trained models for MT evaluation systems, and (3) the potential benefits of an additional MT evaluation dataset in African languages for modeling. 

Our models are based upon the estimator framework~\citep{rei-etal-2020-comet}, as illustrated in Figure~\ref{fig:model_structure} of Appendix~\ref{sec:appendix_a}. In this architecture, the source~(src), machine translation~(mt), and reference translation~(ref) are encoded separately using a multilingual encoder. The resulting word embeddings are passed through a pooling layer to create a sentence embedding for each segment. These sentence embeddings are then combined into a single vector and fed into a feed-forward regressor. The model is trained to minimize the mean squared error. We refer to this as ``\textbf{single-task learning}''~(STL). Furthermore, we adopt a unified approach~\citep{wan2022unite}, which integrates the tasks of <src, mt>, <mt, ref>, and <src, mt, ref> into one model, feeding all three inputs into the pre-trained model and uniformly distributing weight across the three sentence-level scores for the final score prediction for MT evaluation. We refer to this as ``\textbf{multi-task learning}''~(MTL).


\subsection{Experimental Settings}
\subsubsection{Datasets}
\label{sec:dataset}
The adequacy \textbf{dev} sets in \textsc{AfriMTE} are employed as validation sets for modeling purposes, while the adequacy \textbf{devtest} sets serve as the test sets.


Since 2017, organizers of WMT News translation tasks have been gathering human evaluation using the DA method~\citep{graham-etal-2013-continuous}. In addition, another large sourced DA annotation set is the MLQE-PE datasets~\citep{fomicheva2020mlqe}, typically used in WMT Quality Estimation Shared Tasks~\citep{specia-etal-2020-findings-wmt, specia2021findings, zerva2022findings}. We employ these DA datasets as our primary training data,\footnote{We use the DA data from WMT 2017 to 2020 translation tasks and the MLQE-PE data in this work, which can be found in: \url{https://github.com/Unbabel/COMET/tree/master/data}. More information is detailed in~\citet{freitag2021results}.} similar to their application in training the COMET metric~(COMET22)~\citep{rei2022comet}. We label this training data as ``\textbf{WMT Others}''.

Recently, WMT 2022 Large-Scale African Machine Translation Shared Task\footnote{\url{https://www.statmt.org/wmt22/large-scale-multilingual-translation-task.html}} introduces a DA dataset of 99 source sentences from the FLORES-200 test set~\citep{adelani-etal-2022-findings}, covering 46 African language pairs across eight MT engines. Despite its utility, it exhibits two potential limitations: (1) the source context is constrained, consisting of only 99 sentences, and (2) each translation is annotated by a single annotator, raising concerns about the reliability of the assessments. We refer to this dataset as ``\textbf{WMT African}''. 

Statistical summaries of the ``WMT Others'' and ``WMT African'' datasets are provided in Table~\ref{tab:wmt_previous} and Table~\ref{tab:wmt_african} respectively in Appendix~\ref{sec:appendix_a}. Duplicates of <src, mt, ref, DA score> have been excluded. During preprocessing, we also apply z-normalization at the annotator level; to facilitate interpretability and manage the unbounded nature of the quality scores, we apply min-max scaling to the normalized z-scores, adjusting their range to fall between 0 and 1.

\subsubsection{Model configurations} 
\label{sec:model_config}

In the model setup, we utilize three 
multilingual pre-trained encoders: XLM-R-L~\citep{conneau2019unsupervised}, InfoXLM-L~\citep{chi2020infoxlm}, and an XLM-R-L model adapted to 17 African languages---AfroXLM-R-L~\citep{alabi-etal-2022-adapting}. Among these, XLM-R-L and InfoXLM-L have been used in the development of COMET22~\citep{rei2022comet} and CometKiwi~\citep{rei-etal-2022-cometkiwi} metrics for WMT 2022 MT Evaluation and QE Shared Tasks.
We provide a detailed overview of language coverage of these three models in Table~\ref{tab:language coverage} of Appendix~\ref{sec:appendix_a}.

We train our models with the open-source COMET codebase.\footnote{\url{https://github.com/Unbabel/COMET}} Training of each model is executed on a single NVIDIA A100-SXM4-80GB graphics card, with a configured batch size of 16 and a gradient accumulation across 2 batches. We follow the default settings for other hyper-parameters of the COMET metric.\footnote{Hyper-parameters are configured at~\url{https://github.com/Unbabel/COMET/tree/master/configs}.}

\subsubsection{Evaluation}
Pearson, Spearman-rank, and Kendall-rank are widely-used correlation coefficients to assess the correlation between automated and human-annotated scores. Recent findings~\citep{deutsch2023ties} indicate that Pearson is complementary to Kendall, and Spearman balances between Pearson's effectiveness in noisy but linear scenarios and Kendall's in ordered but non-linear ones. Thus, we utilize the Spearman-rank correlation coefficient as our primary monitoring metric during model training. For testing, we report all 3 coefficients. To validate the statistical significance of our results, we employ the Perm-Input hypothesis test~\citep{deutsch2021statistical}, conducting 200 re-sampling runs and setting $p = 0.05$. It produces rankings of various automatic metrics. Essentially, for a given test set (in our case, this encompasses translations and their respective assessments), two metrics are juxtaposed using diverse subsets derived from the original test data. The final ranks stem from a significance matrix, which comprises comparisons between all possible pairs of metrics.

\input{tables/spearman_main}

\subsection{Main Findings}
In this section, we will present our experimental results for our investigations around the three key questions mentioned at the beginning of Section~\ref{sec:afriCOMET}.

\subsubsection{Transfer learning from well-resourced DA data with pre-trained encoders}
Initially, we develop our MT evaluation systems that leverage transfer learning from a variety of well-resourced languages to African languages. We train our models on ``WMT Others'' and employ the adequacy dev and devtest sets within~\textsc{AfriMTE} as validation and test sets. As outlined in Section~\ref{sec:model_config}, to explore the impact of various multilingual encoders, we conduct experiments based on XLM-R-L, InfoXLM-L, and AfroXLM-R-L for comparison. In our comparison, we benchmark our models against (1) the widely used n-gram matching based evaluation metrics SacreBLEU~\citep{post-2018-call} and chrf++~\citep{popovic-2017-chrf}, (2) the embedding-based metric, BERTScore~\citep{Zhang_020BERTScore}, (3) LLM Prompting based GPT-4 output with OpenAI API\footnote{We use the ``gpt-4-0613'' version, prompting it with the meta-prompt outlined in Figure~\ref{fig:gpt4-prompt} of Appendix~\ref{sec:appendix_a}.} and (4) the learned COMET22 metric~\citep{rei2022comet}, which uses the XLM-R-L encoder and also ``WMT Others'' as training data, but differs in validation, employing additional MQM data for English-German, Chinese-English, and English-Russian from the WMT 2022 News Shared Task.\footnote{\url{https://github.com/google/wmt-mqm-human-evaluation}}

Results of sentence-level Spearman-rank correlation coefficients are shown in Table~\ref{tab:spearman_main_result}. Given that ``WMT Others'' does not include any African language except English, the results of ``Learned COMET Metrics'' illuminate the effectiveness of various pre-trained multilingual encoders for zero-shot scenarios. 
Among them, AfroXLM-R-L achieves the highest average result, demonstrating a promising ability to transfer knowledge from well-resourced languages to under-resourced African languages with an African enhanced multilingual encoder. Its performance is further improved with ``multi-task learning''. We also present Pearson and Kendall-rank correlation coefficient results in Table~\ref{tab:pearson_kendall_main_result} in Appendix~\ref{sec:appendix_a}, and the trends observed are consistent with those derived from the Spearman's analysis. Results of Perm-Input hypothesis test for 3 coefficients are illustrated in Table~\ref{tab:spearman_main_rank}, \ref{tab:pearson_main_rank} and~\ref{tab:kendall_main_rank} respectively in Appendix~\ref{sec:appendix_a}. Both AfroXLM-R-L based systems~(STL and MTL) tend to outperform N-gram matching based metrics, BERTScore and COMET22, and show comparable or superior results to GPT-4. 

Particularly, our results reveal improvements for eng-ibo and eng-yor~(FLORES, News, and TED talks) when we utilize AfroXLM-R-L instead of XLM-R-L as encoder, aligning with their language coverage in Table~\ref{tab:language coverage} in Appendix~\ref{sec:appendix_a}.
Additionally, languages initially supported by XLM-R-L, such as eng-hau, eng-som and eng-xho, also experience enhancements with the adoption of AfroXLM-R-L.
Interestingly, eng-kik and eng-luo translation evaluations show marked improvements even though Kikuyu and Luo are not explicitly covered by AfroXLM-R-L.
Further analysis of correlations across four eng-yor domain-specific datasets show that models trained based on AfroXLM-R-L have the potential to surpass the performance of COMET22, indicating its generalization for different domains despite being trained on the News, Wikipedia and Health domains. 
For the control LP, eng-fra, our AfroXLM-R-L based systems are among the top-performing systems, with distinctions underscored by their bolded rankings. 
Notably, GPT-4 shows impressive performance in eng-yor and yor-eng MT evaluations.

\subsubsection{Impact of an extra African DA dataset}
To discuss the potential benefits of an additional MT evaluation dataset in African languages, we conduct experiments based on AfroXLM-R-L across three distinct training data configurations: (1) \textbf{``WMT African''}, (2) \textbf{``WMT Others''}, and (3) a merged dataset of \textbf{``WMT African''} and \textbf{``WMT Others''}, which we refer to as \textbf{``WMT Combined''}. The STL and MTL results, including Pearson, Spearman-rank, Kendall-rank correlation coefficients, and Perm-Input hypothesis test results, are detailed in Table~\ref{tab:afr_xlmr_result}, \ref{tab:afr_xlmr_stl_rank}, \ref{tab:afr_xlmr_result_mlt} and \ref{tab:afr_xlmr_mtl_rank} respectively in Appendix~\ref{sec:appendix_a}. Remarkably, ``WMT Others'' yields higher Spearman-rank and Kendall-rank correlations than ``WMT Combined''. While ``WMT Combined'' shows the highest Pearson correlation, it negatively impacts both Spearman-rank and Kendall-rank correlations. Examining all three correlation coefficients and the Perm-Input hypothesis test results reveals that models trained on ``WMT Others'' and ``WMT Combined'' significantly outperform the model trained solely on ``WMT African''. This disparity in performance could be attributed to the limited size and diversity in the context of ``WMT African'', suggesting its data scarcity issue. In summary, leveraging transfer learning from ``WMT Others'' based on AfroXLM-R-L proves effective in building African COMET models.
\input{tables/ref_free_result}

\subsection{The benchmark MT evaluation metrics}
The AfroXLM-R-L based STL and MTL models, marked with \ding{72} in Table~\ref{tab:spearman_main_result}, are established as our benchmark MT evaluation systems for African languages. They achieve a Spearman-rank correlation up to 0.441 with human judgments and are named with AfriCOMET-STL and AfriCOMET-MTL.



%% file: tables/spearman_main.tex
\begin{table*}[!t]
\begin{center}
\centering
\scalebox{0.6}{
\begin{tabular}{l|cc|c|c|c|ccc|c}
\toprule
& \multicolumn{2}{c|}{\textbf{N-gram Matching}} 
& \multicolumn{1}{c|}{\textbf{Embedding-based}} 
& \multicolumn{1}{c|}{\textbf{LLM Prompting}}
& \multicolumn{5}{c}{\textbf{Learned COMET Metrics}} \\
\cmidrule{2-10}
& \multicolumn{1}{c}{\multirow{2}{*}{SacreBLEU}} 
& \multicolumn{1}{c|}{\multirow{2}{*}{chrf++}}
& \multicolumn{1}{c|}{\multirow{2}{*}{BERTScore}}
& \multicolumn{1}{c|}{\multirow{2}{*}{GPT-4}} 
& \textbf{Baseline} 
& \multicolumn{3}{c|}{\textbf{Single Task} (Ours)} 
& \textbf{Multi Task} (Ours) \\
\cmidrule{6-10}
LP & & & & & COMET22 & XLM-R-L & InfoXLM-L & AfroXLM-R-L \ding{72} & AfroXLM-R-L \ding{72} \\
\midrule
ary-fra	&	0.332	&	0.328	&	0.351	&	\textbf{0.620}	&	0.533	&	0.551	&	\textbf{0.565}	&	\textbf{0.567}	&	\textbf{0.609}	\\
eng-arz	&	0.324	&	0.321	&	0.355	&	0.509	&	0.503	&	0.486	&	0.488	&	0.532	&	\textbf{0.600}	\\
eng-fra	&	0.246	&	0.280	&	0.282	&	\textbf{0.536}	&	\textbf{0.489}	&	\textbf{0.510}	&	0.460	&	\textbf{0.495}	&	\textbf{0.526}	\\
eng-hau	&	0.200	&	0.301	&	0.404	&	0.378	&	0.430	&	0.401	&	0.334	&	0.515	&	\textbf{0.620}	\\
eng-ibo	&	0.339	&	0.424	&	0.403	&	0.271	&	0.373	&	0.413	&	0.377	&	\textbf{0.592}	&	\textbf{0.616}	\\
eng-kik	&	0.273	&	\textbf{0.295}	&	0.276	&	0.269	&	0.202	&	0.281	&	0.249	&	\textbf{0.389}	&	\textbf{0.410}	\\
eng-luo	&	0.182	&	0.279	&	\textbf{0.365}	&	0.246	&	0.062	&	0.201	&	0.241	&	\textbf{0.283}	&	\textbf{0.359}	\\
eng-som	&	0.161	&	0.279	&	0.345	&	0.281	&	0.474	&	0.466	&	0.420	&	\textbf{0.554}	&	\textbf{0.546}	\\
eng-swh	&	0.481	&	0.565	&	0.701	&	\textbf{0.774}	&	\textbf{0.738}	&	\textbf{0.739}	&	\textbf{0.719}	&	0.688	&	\textbf{0.733}	\\
eng-twi	&	\textbf{0.204}	&	\textbf{0.178}	&	0.111	&	\textbf{0.132}	&	0.096	&	0.103	&	\textbf{0.112}	&	\textbf{0.157}	&	0.101	\\
eng-xho	&	0.090	&	\textbf{0.161}	&	\textbf{0.168}	&	\textbf{0.143}	&	0.071	&	0.070	&	0.059	&	\textbf{0.191}	&	\textbf{0.146}	\\
eng-yor	&	0.210	&	0.204	&	0.250	&	\textbf{0.446}	&	0.150	&	0.193	&	0.191	&	0.287	&	0.365	\\
eng-yor (it)	&	0.295	&	0.346	&	\textbf{0.421}	&	\textbf{0.447}	&	0.334	&	0.256	&	0.268	&	0.266	&	\textbf{0.418}	\\
eng-yor (movie)	&	0.238	&	0.221	&	0.303	&	\textbf{0.544}	&	0.334	&	0.338	&	0.364	&	0.372	&	0.390	\\
eng-yor (news)	&	0.114	&	0.122	&	0.111	&	\textbf{0.200}	&	\textbf{0.168}	&	\textbf{0.196}	&	\textbf{0.132}	&	\textbf{0.200}	&	\textbf{0.211}	\\
eng-yor (ted)	&	0.027	&	0.002	&	0.091	&	\textbf{0.237}	&	0.123	&	0.177	&	\textbf{0.263}	&	\textbf{0.324}	&	\textbf{0.298}	\\
yor-eng	&	0.308	&	0.408	&	0.446	&	\textbf{0.476}	&	\textbf{0.502}	&	0.460	&	\textbf{0.481}	&	\textbf{0.490}	&	\textbf{0.541}	\\
\midrule
Avg.	&	0.237	&	0.277	&	0.317	&	0.383	&	0.328	&	0.344	&	0.337	&	\textbf{0.406}	&	\textbf{0.441}	\\
\bottomrule
\end{tabular}
}
\end{center}
\caption{Sentence-level Spearman-rank correlation coefficients for MT evaluation models.
For each LP, values in~\textbf{bold} represent the highest ranking obtained from the Perm-Input hypothesis test~\citep{deutsch2021statistical}. Comprehensive results of this test are detailed in Table~\ref{tab:spearman_main_rank}. Averaged Spearman-rank correlations across LPs are presented in the last row.}
\label{tab:spearman_main_result}
\vspace{-0.5cm}
\end{table*}

%% file: tables/ref_free_result.tex
\begin{table*}[!t]
\centering
\resizebox{2.0\columnwidth}{!}{%
\begin{tabular}{l|cc|cc|cc|cc|cc}
\toprule
& \multicolumn{2}{c|}{\textbf{LLM Prompting}} 
& \multicolumn{8}{c}{\textbf{Learned reference-free QE Metrics}} \\
\cmidrule{2-11}
& \multicolumn{2}{c|}{\multirow{2}{*}{GPT-4}} 
& \multicolumn{2}{c|}{\textbf{Baseline}}
& \multicolumn{4}{c|}{\textbf{Single Task} (Ours)}
& \multicolumn{2}{c}{\textbf{Multi Task} (Ours)} \\
\cmidrule{4-11}
& & & \multicolumn{2}{c|}{CometKiwi}
& \multicolumn{2}{c|}{InfoXLM-L}
& \multicolumn{2}{c|}{AfroXLM-R-L \ding{72}}
& \multicolumn{2}{c}{AfroXLM-R-L \ding{72}}  \\
\cmidrule{2-11}
LP & Pearson & Spearman & Pearson & Spearman & Pearson & Spearman & Pearson & Spearman & Pearson & Spearman \\
\midrule
ary-fra	&	\textbf{0.660}	&	\textbf{0.622}	&	0.517	&	0.495	&	0.530	&	\textbf{0.561}	&	0.475	&	0.507	&	\textbf{0.610}	&	\textbf{0.534}	\\
eng-arz	&	0.462	&	\textbf{0.525}	&	\textbf{0.611}	&	\textbf{0.592}	&	0.562	&	0.516	&	0.551	&	0.516	&	\textbf{0.600}	&	\textbf{0.580}	\\
eng-fra	&	\textbf{0.471}	&	\textbf{0.531}	&	\textbf{0.527}	&	\textbf{0.495}	&	0.416	&	\textbf{0.484}	&	\textbf{0.418}	&	\textbf{0.478}	&	\textbf{0.483}	&	\textbf{0.531}	\\
eng-hau	&	0.363	&	0.284	&	0.314	&	0.245	&	0.382	&	0.273	&	\textbf{0.652}	&	0.482	&	\textbf{0.690}	&	\textbf{0.586}	\\
eng-ibo	&	0.175	&	0.105	&	0.205	&	0.188	&	0.335	&	0.334	&	\textbf{0.644}	&	\textbf{0.631}	&	\textbf{0.597}	&	\textbf{0.574}	\\
eng-kik	&	0.144	&	0.198	&	0.277	&	0.247	&	0.409	&	\textbf{0.339}	&	\textbf{0.631}	&	\textbf{0.415}	&	0.437	&	0.317	\\
eng-luo	&	0.038	&	-0.044	&	0.237	&	\textbf{0.161}	&	0.142	&	\textbf{0.130}	&	\textbf{0.333}	&	\textbf{0.217}	&	0.256	&	\textbf{0.174}	\\
eng-som	&	0.179	&	0.219	&	\textbf{0.266}	&	0.357	&	0.155	&	0.251	&	\textbf{0.302}	&	\textbf{0.482}	&	\textbf{0.302}	&	\textbf{0.510}	\\
eng-swh	&	0.693	&	\textbf{0.731}	&	\textbf{0.787}	&	\textbf{0.756}	&	0.699	&	0.637	&	0.644	&	0.587	&	0.737	&	\textbf{0.718}	\\
eng-twi	&	\textbf{0.212}	&	\textbf{0.053}	&	0.097	&	\textbf{0.026}	&	-0.003	&	-0.050	&	\textbf{0.290}	&	\textbf{0.061}	&	\textbf{0.279}	&	\textbf{0.060}	\\
eng-xho	&	\textbf{0.254}	&	\textbf{0.119}	&	0.127	&	-0.030	&	0.190	&	\textbf{0.041}	&	\textbf{0.437}	&	\textbf{0.085}	&	\textbf{0.472}	&	\textbf{0.130}	\\
eng-yor	&	0.339	&	\textbf{0.357}	&	0.327	&	0.231	&	0.489	&	0.225	&	\textbf{0.738}	&	\textbf{0.392}	&	0.643	&	0.280	\\
eng-yor (it)	&	0.308	&	0.283	&	0.375	&	\textbf{0.388}	&	0.299	&	0.304	&	\textbf{0.654}	&	0.318	&	\textbf{0.641}	&	\textbf{0.419}	\\
eng-yor (movie)	&	0.411	&	\textbf{0.472}	&	0.151	&	0.041	&	0.328	&	0.240	&	\textbf{0.557}	&	0.314	&	0.450	&	0.311	\\
eng-yor (news)	&	0.239	&	\textbf{0.126}	&	0.104	&	0.078	&	0.219	&	0.057	&	\textbf{0.508}	&	\textbf{0.186}	&	\textbf{0.496}	&	\textbf{0.206}	\\
eng-yor (ted)	&	\textbf{0.310}	&	\textbf{0.246}	&	0.217	&	\textbf{0.289}	&	0.267	&	\textbf{0.218}	&	\textbf{0.518}	&	\textbf{0.189}	&	\textbf{0.409}	&	\textbf{0.271}	\\
yor-eng	&	\textbf{0.383}	&	\textbf{0.399}	&	0.070	&	0.098	&	-0.007	&	0.059	&	0.181	&	0.208	&	\textbf{0.383}	&	\textbf{0.414}	\\
\midrule
Avg.	&	0.332	&	0.307	&	0.306	&	0.274	&	0.318	&	0.272	&	\textbf{0.502}	&	\textbf{0.357}	&	\textbf{0.499}	&	\textbf{0.389}	\\
\bottomrule
\end{tabular}
}
\caption{Sentence-level correlation coefficients (Pearson, Spearman-rank) for QE models. 
For each LP, values in~\textbf{bold} represent the highest ranking obtained from the Perm-Input hypothesis test~\citep{deutsch2021statistical}. The comprehensive results of this test are detailed in Table~\ref{tab:ref_free_rank}. 
Averaged correlations across LPs are presented in the last row.
}
\label{tab:ref_free_result}
\vspace{-0.5cm}
\end{table*}

%% file: ref_free_model.tex
\section{Reference-free QE systems}
Utilizing adequacy annotations within \textsc{AfriMTE}, we are also able to develop reference-free models that predict translation quality in the absence of reference translations, aligning with research advancements in translation quality estimation~(QE)~\citep{fan2019bilingual, ranasinghe2020transquest, specia-etal-2020-findings-wmt, wang2021beyond, wang2021qemind, specia2021findings, rei-etal-2022-cometkiwi, zerva2022findings}. Our QE models adhere to the same Estimator architecture as AfriCOMET, but excluding the reference translation from model inputs. Both STL and MTL methods can be applied. However, different from applying MTL in MT evaluation, once the multi-task model is trained, it strictly requires <src, mt> as the input for inference and only generates the corresponding <src, mt> score as its final score.

We choose AfroXLM-R-L and InfoXLM-L for comparison and train our QE models on ``WMT Others''.\footnote{We follow the hyper-parameter settings at~\url{https://github.com/Unbabel/COMET/tree/master/configs}, use the same batch size and gradient accumulation, and utilize the same hardware as when training the MT evaluation models.} These models are validated and evaluated using adequacy dev and devtest sets within \textsc{AfriMTE}. We benchmark our QE systems against Prompting GPT-4 with meta-prompt as shown in Figure~\ref{fig:gpt4-prompt} of Appendix~\ref{sec:appendix_a} and CometKiwi~\citep{rei-etal-2022-cometkiwi}, which is trained on ``WMT Others'' and leverages InfoXLM-L as its encoder. 
%
%

When prompting GPT-4, we encounter certain challenges where the API sometimes fails to return the anticipated quality scores, likely attributable to two reasons: the inherent unpredictability of GPT-4's generative capability and its difficulty in identifying some extremely low-resource languages such as Kikuyu, Igbo and Twi in the QE scenario.\footnote{We re-query the GPT-4 API up to five times for each example, in an effort to obtain successful responses. Despite these efforts, certain instances persist where error responses are encountered even after five attempts.} We find that for eng-kik, eng-ibo, and eng-twi, error rates are markedly higher than the general trend, recorded at 26.2\%, 7.5\%, and 11.7\%, compared to an overall error occurrence below 5\% for other LPs.
%
%
%
Therefore, for each LP, we implement a missing data imputation approach by assigning the mean of outputs from the remaining valid examples to those that get error responses to ensure consistency and fairness in our evaluation.


QE systems are commonly assessed using Pearson and Spearman-rank correlations as highlighted in~\citep{zerva2022findings}. Our results, showcased in Table~\ref{tab:ref_free_result}, along with the Perm-Input hypothesis test results in Table~\ref{tab:ref_free_rank} in Appendix~\ref{sec:appendix_a}, reveal the following insights. The InfoXLM-L STL model, trained on ``WMT Others'', performs on par with CometKiwi under the same encoder configurations. However, the AfroXLM-R-L STL model exhibits significant improvements in both Pearson and Spearman-rank correlations, superior over CometKiwi. Additionally, MTL training further boosts performance in Spearman-rank correlation. These highlight the effectiveness of transfer learning from robust, well-resourced DA data, especially when utilizing AfroXLM-R-L as the pre-trained encoder for the reference-free QE task. 

Moreover, when we compare Spearman-rank results in Table~\ref{tab:spearman_main_result} and~\ref{tab:ref_free_result}, AfroXLM-R-L based QE systems~(STL and MTL) outperform GPT-4 by a larger margin than observed in MT evaluation, and the performance gap between QE and MT evaluation systems is larger with GPT-4 $(0.076=0.383 - 0.307)$ compared to the AfroXLM-R-L based systems, $(0.049=0.406 - 0.357)$ for STL and $(0.052=0.441 - 0.389)$ for MTL. This highlights GPT-4's challenges with QE tasks and underscores the superior efficacy of our supervised systems in addressing the inherently cross-lingual nature of QE, diverging from the MT evaluation task. The latter typically involves easier monolingual pattern-matching tasks in comparing machine translations against reference translations.

Finally, we introduce our benchmark QE systems for African MT: the AfroXLM-R-L based STL and MTL models marked with \ding{72} in Table~\ref{tab:ref_free_result}, and name them with AfriCOMET-QE-STL and AfriCOMET-QE-MTL.\footnote{Please note that AfriCOMET-QE-MTL and AfriCOMET-MTL are identical in training, as both are trained using the same multi-task learning approach.}

%% file: paper_limitation.tex
\section*{Limitations}
This work establishes an efficient solution to translation evaluation for under-resourced African languages. It shows that with leveraging a pre-trained model enhanced by under-resourced languages, it is feasible to transfer knowledge from well-resourced to under-resourced languages for the downstream cross-lingual NLP tasks. However, our current methods are subject to limitations. 

Firstly, while using AfroXLM-R-L as a pre-trained encoder enhances the performance of our benchmark systems for certain language pairs such as eng-ibo, eng-kik, eng-luo and eng-yor, this improvement isn't consistent across all LPs. For example, the eng-twi translation evaluation shows no such enhancement and Twi is also not covered by AfroXLM-R-L. Addressing the limited resources and coverage for such under-resourced languages remains a challenge for future work.

Secondly, both our MT evaluation and QE benchmark systems are developed using adequacy annotations within~\textsc{AfriMTE}, mainly drawing inspiration from works by~\citet{bentivogli2018machine, chatzikoumi2020evaluate}, which suggest that overall DA largely focuses on adequacy. However, upon analyzing the correlations between adequacy and fluency annotations, we have observed a slight negative correlation between total fluency error counts in a translation and its adequacy DA score, with a Pearson correlation coefficient of $-0.349$. This raises a question: \textit{whether incorporating fluency assessments in developing MT evaluation and QE models could yield any benefit?} Exploring this possibility will be another area for future work.

Thirdly, comparisons of the Spearman-rank results in Table~\ref{tab:spearman_main_result} and~\ref{tab:ref_free_result} show significant performance gaps between the AfroXLM-R-L based MT evaluation and reference-free QE systems, even though both employing transfer learning. This disparity likely arises from the tasks' different natures: MT evaluation models are trained with the help of reference inputs, resembling monolingual pattern-recognition tasks that compare machine translations with references. However, the QE task, inherently cross-lingual due to its reference-free nature, highlights the potential need for more training data to bridge this gap. This will be another focus in our future research.

Fourthly, the test datasets in this study, currently limited to translations from a single MT engine per LP, could benefit from diversification. Incorporating outputs from various MT systems into our annotated test data would enrich the spectrum of MT errors observed, significantly enhancing the robustness of metric evaluations. This would be particularly advantageous for developing ranking systems for translation evaluation. However, the expansion is constrained by limited annotation resources, a challenge that is more pronounced for under-resourced African languages within the context of our simplified MQM approach. Despite these challenges, this work's primary goal is to establish reliable metrics for assessing sentence-level translation quality for under-resourced African languages. Our findings demonstrate that our benchmark systems can be used to assess translations from various translation engines.

\section*{Ethics Statement}
Our work and collection of data has been deeply rooted in the principles of participatory AI research~\citep{nekoto-etal-2020-participatory}, where the native speakers, most affected by lack of evaluation metrics, are involved throughout the project as stakeholders. They contributed to the data and gave their consent to use this data for the enhancement of COMET models for African languages. Upon the data collected, there is no privacy concern since the source of the data is based on Wikipedia general domain.

%% file: sec-app.tex
\section{Appendix}
\label{sec:appendix}

\subsection{Supplementary materials}
\label{sec:appendix_a}
\begin{figure*}[h]
\centering
\begin{subfigure}{0.98\linewidth}
  \centering
\includegraphics[width=0.95\textwidth]{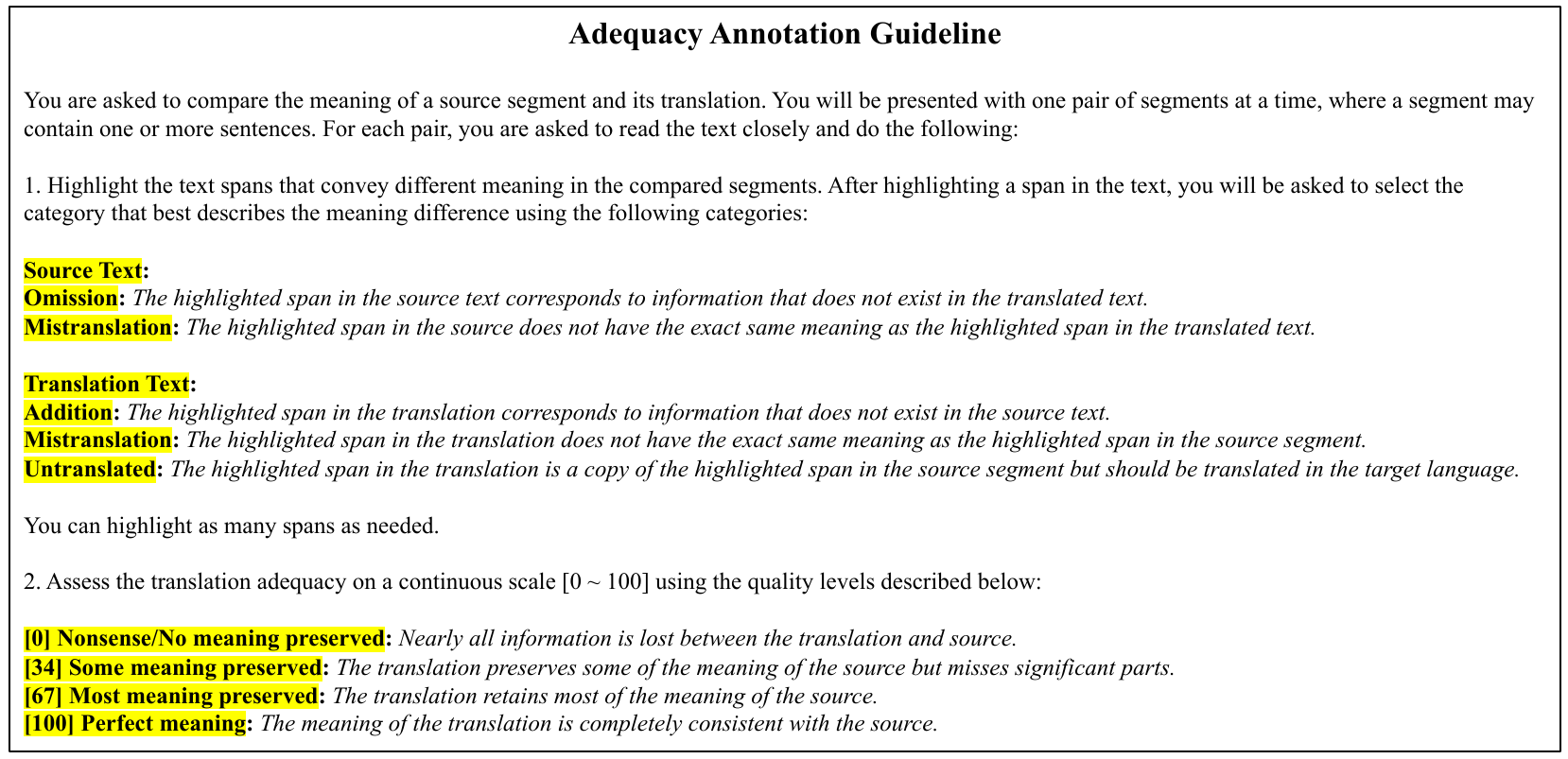}
\end{subfigure}%
\caption{\textbf{Adequacy annotation guideline} for error highlighting [the first part] and DA score assignment [the second part].} 
\label{fig:ade_guideline}
\end{figure*}

\begin{figure*}[h]
\centering
\begin{subfigure}{0.98\linewidth}
  \centering
\includegraphics[width=0.95\textwidth]{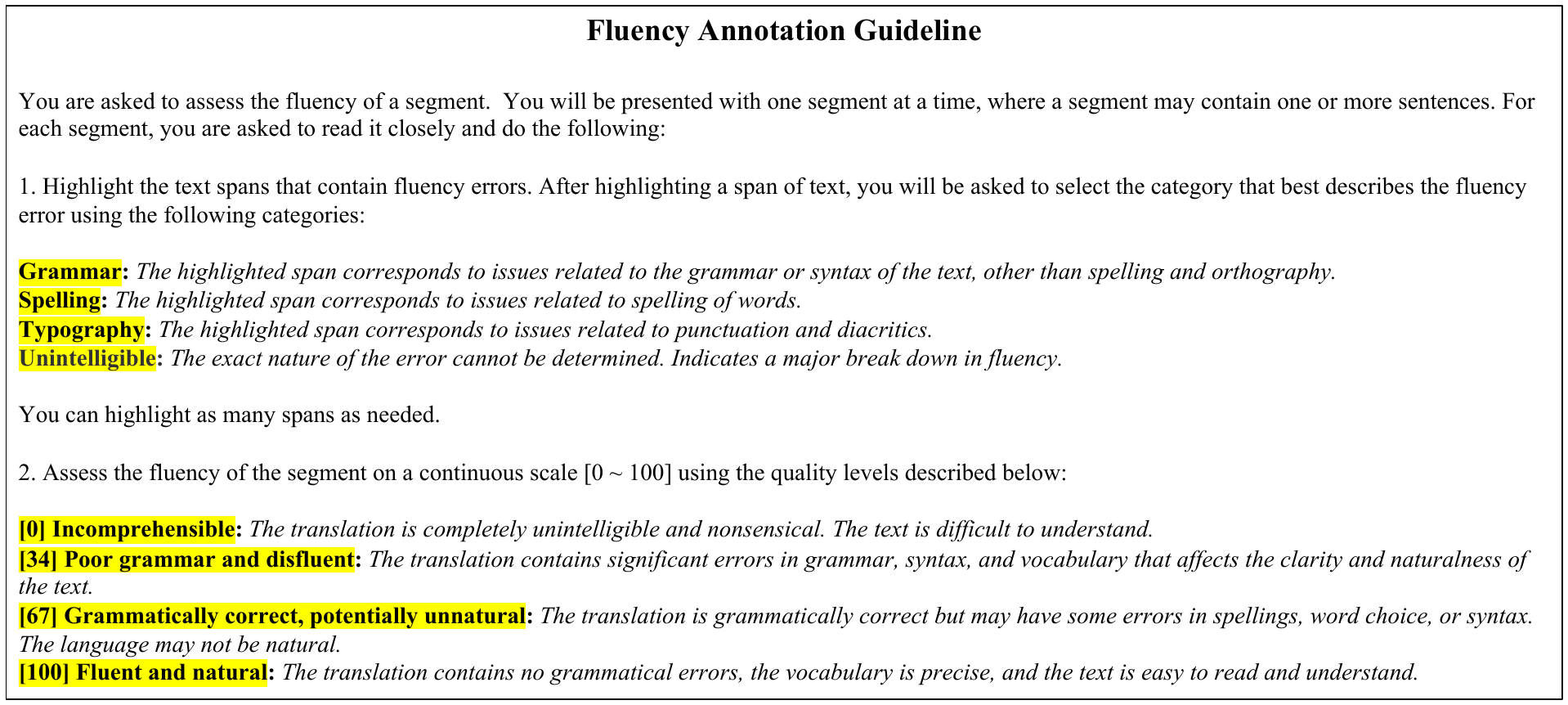}
\end{subfigure}%
\caption{\textbf{Fluency annotation guideline} for error highlighting [the first part] and DA score assignment [the second part].} 
\label{fig:flu_guideline}
\end{figure*}

\begin{figure*}[h]
\centering
\begin{subfigure}{2.0\columnwidth}
\centering
\includegraphics[width=1.0\textwidth]{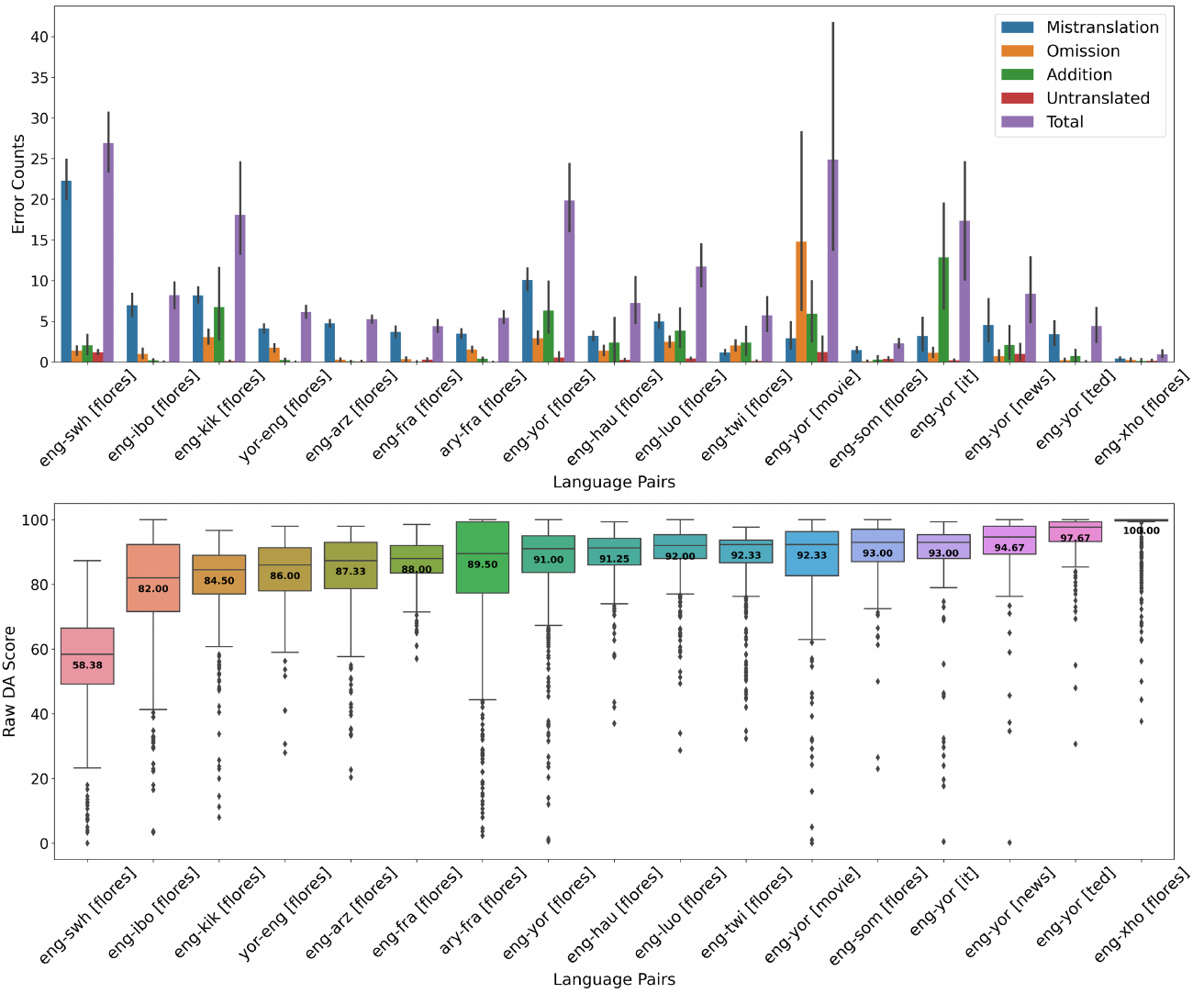}
\end{subfigure}%
\caption{Counts of each error category and sentence-level translation quality measured by DA scores across all language pairs and domains for~\textbf{adequacy}.} 
\label{fig:span_plot}
\end{figure*}

\begin{figure*}[h]
\centering
\begin{subfigure}{2.0\columnwidth}
\centering
\includegraphics[width=1.0\textwidth]{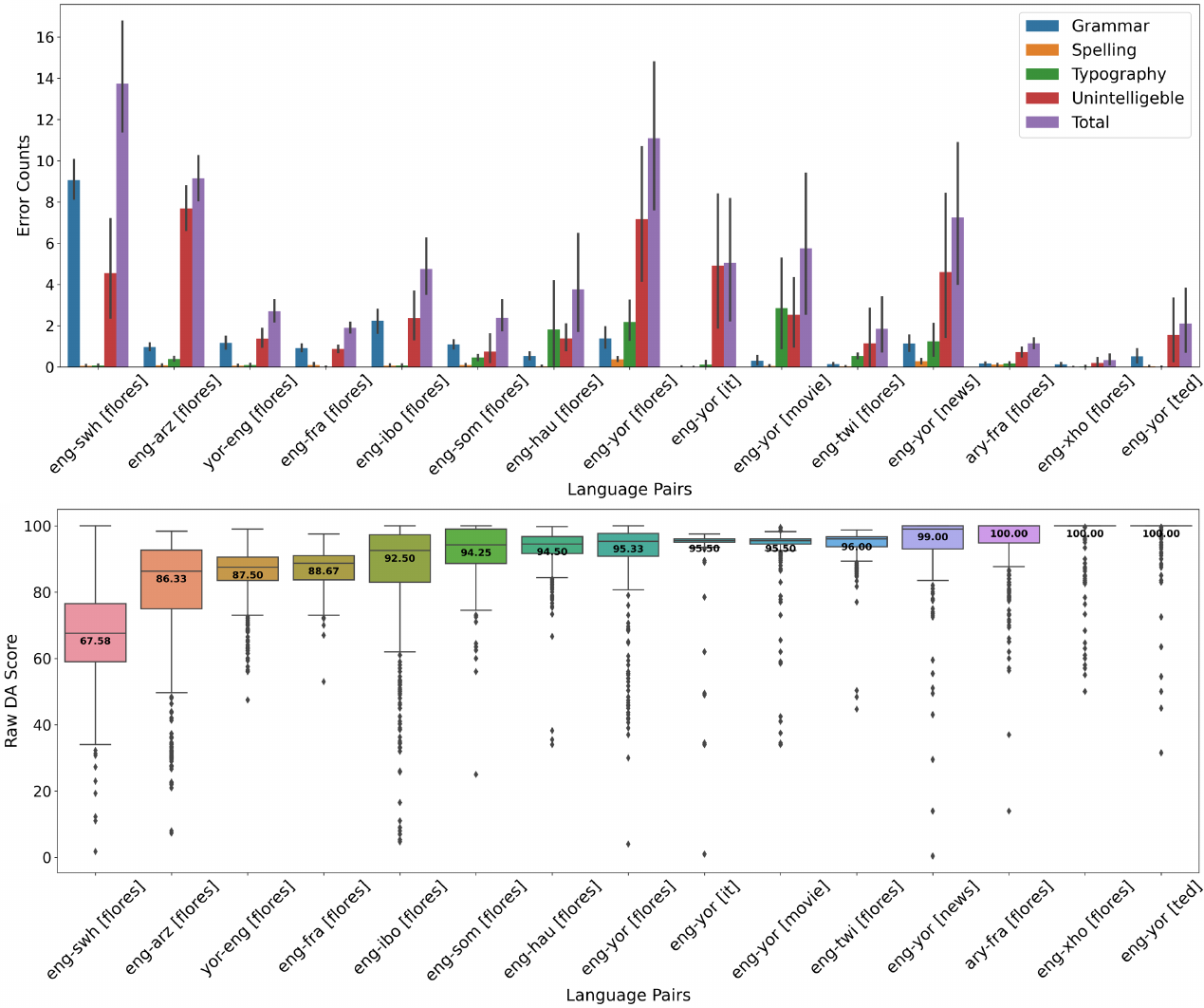}
\end{subfigure}%
\caption{Counts of each error category and sentence-level translation quality measured by DA scores across all language pairs and domains for~\textbf{fluency}.} 
\label{fig:span_plot_flu}
\end{figure*}

\begin{figure*}[h]
\centering
\begin{subfigure}{1.0\columnwidth}
\centering
\includegraphics[width=0.9\textwidth]{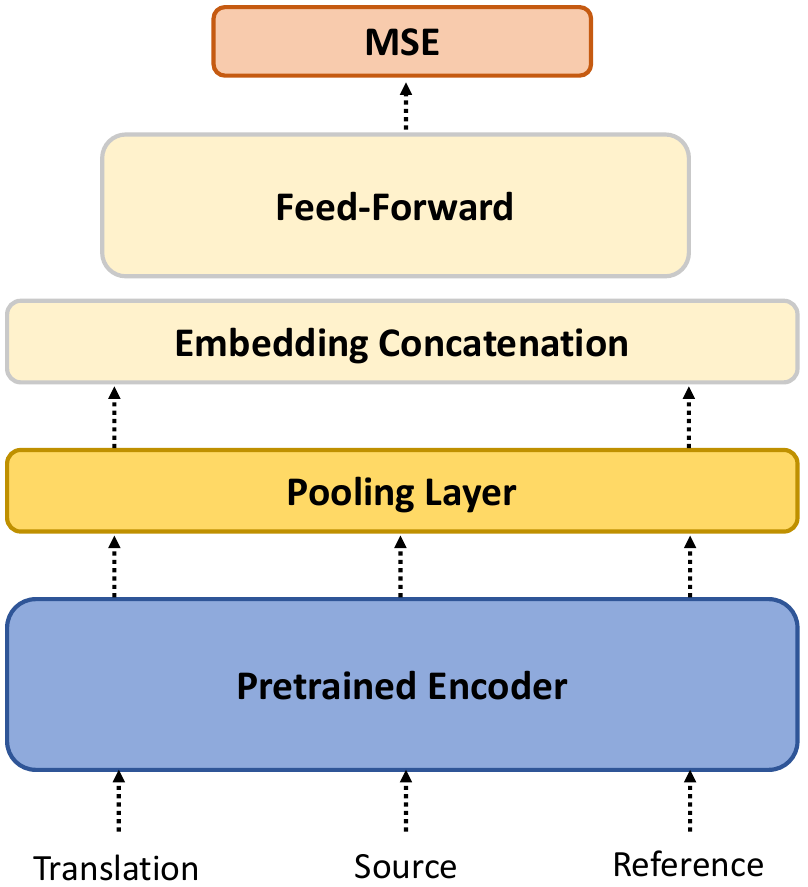}
\end{subfigure}%
\caption{Estimator model architecture. A pre-trained cross-lingual encoder independently encodes the source, translation, and reference. The resulting word embeddings are then passed through a pooling layer to create a sentence embedding for each segment. Then, the corresponding sentence embeddings are combined and concatenated into one single vector, passed to a feed-forward regressor. The entire model is trained by minimizing the Mean Squared Error. Please note that only the source and translation are fed into the pre-trained encoder for training a reference-free QE model.} 
\label{fig:model_structure}
\end{figure*}

\begin{figure*}[h]
\centering
\begin{subfigure}{2.0\columnwidth}
\centering
\includegraphics[width=1.0\textwidth]{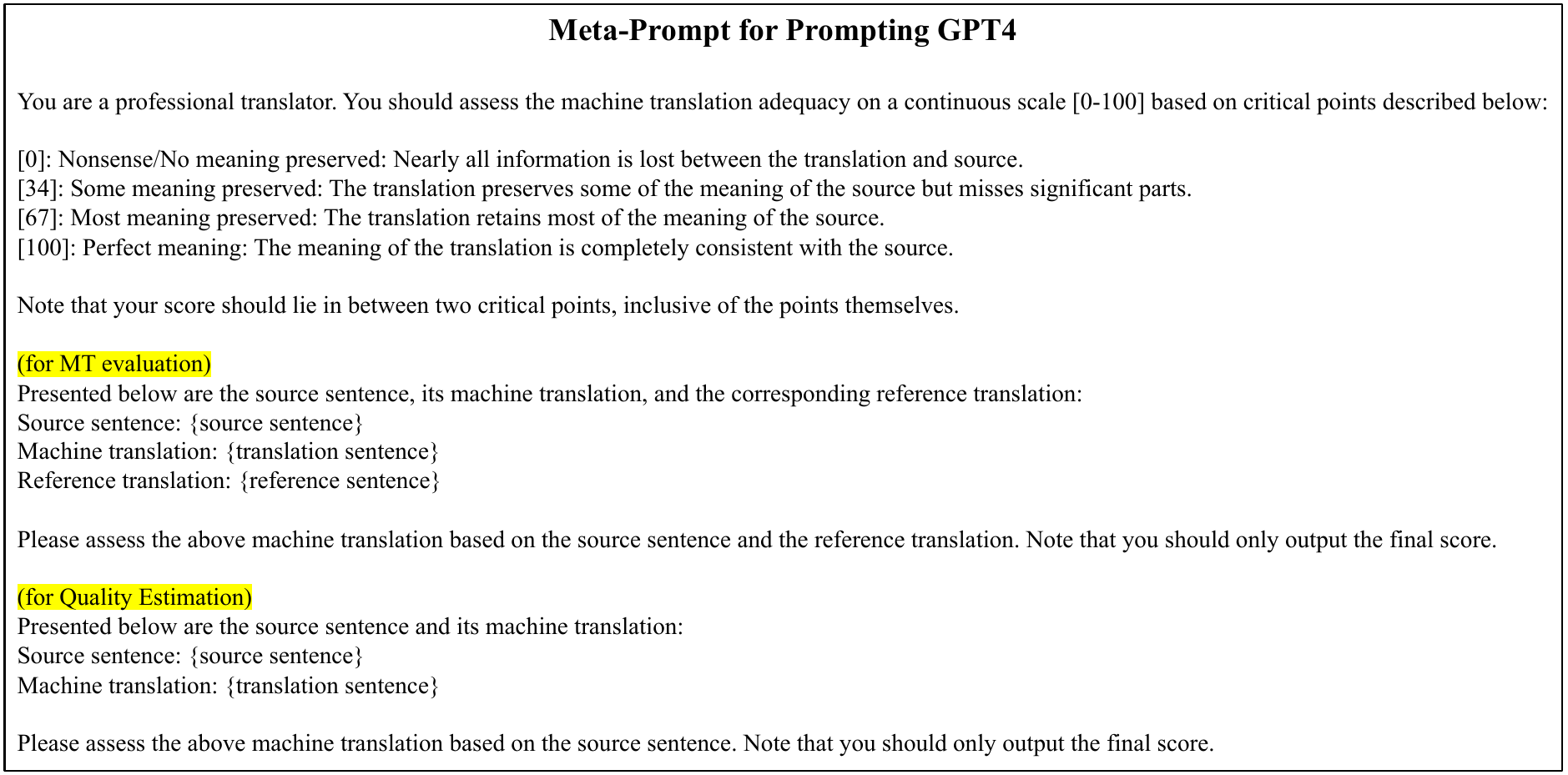}
\end{subfigure}%
\caption{Meta prompts utilized in prompting GPT-4 (version: ``gpt-4-0613'') for MT evaluation and Quality Estimation tasks. Highlights are excluded from the prompts.} 
\label{fig:gpt4-prompt}
\end{figure*}

The Appendix provides supplementary materials supporting the main paper, including 
(i) the information of language family groups that our targeted African languages belong to (Table \ref{tab:language_family}) and the statistics of \textsc{AfriMTE} annotations (Tables \ref{tab:overall_count} and \ref{tab:overall_count_fluency}), 
(ii) detailed simplified annotation guidelines (Figures \ref{fig:ade_guideline} and \ref{fig:flu_guideline}), 
(iii) distributions of error counts and overall sentence-level DA scores of \textsc{AfriMTE} annotations (Figures \ref{fig:span_plot} and \ref{fig:span_plot_flu}), 
(iv) the MT evaluation and QE model architecture (Figure \ref{fig:model_structure}), 
(v) meta-prompts for prompting GPT-4 for MT evaluation and QE tasks (Figure \ref{fig:gpt4-prompt}), 
(vi) statistical summaries of the ``WMT Others'' and ``WMT African'' datasets (Tables \ref{tab:wmt_previous} and \ref{tab:wmt_african}), 
(vii) the overview of language coverage in various pre-trained multilingual models (Table \ref{tab:language coverage}), 
(viii) the Pearson and Kendall-rank correlation coefficients and the Perm-Input hypothesis test results for MT evaluation models (Tables \ref{tab:pearson_kendall_main_result}, \ref{tab:spearman_main_rank}, \ref{tab:pearson_main_rank}, and \ref{tab:kendall_main_rank}), 
(ix) ablation study results for an extra African DA training dataset (Tables \ref{tab:afr_xlmr_result}, \ref{tab:afr_xlmr_stl_rank}, \ref{tab:afr_xlmr_result_mlt}, and \ref{tab:afr_xlmr_mtl_rank}), 
and (x) the Perm-Input hypothesis test results for QE models (Table \ref{tab:ref_free_rank}).

\input{tables/lang_family}
\input{tables/overall_count}
\input{tables/overall_count_fluency}

\input{tables/wmt_previous}

\input{tables/wmt_african}

\input{tables/language_coverage}

\input{tables/pearson_kendall_main}
\input{tables/spearman_main_rank}

\input{tables/afr_xlmr_result}
\input{tables/afr_xlmr_stl_rank}
\input{tables/afr_xlmr_result_mtl}
\input{tables/afr_xlmr_mtl_rank}
\input{tables/ref_free_rank_result}

\subsection{Evaluation on the WMT African DA dataset}
\label{subsec:appendix b}
Besides evaluating AfriCOMET and AfriCOMET-QE using the adequacy devtest sets within \textsc{AfriMTE}, we conduct additional assessments on the ``WMT African'' dataset, despite its potential limitations discussed in Section~\ref{sec:dataset}. These assessments are justified because the ``WMT African'' dataset is not utilized in the development (training or validation) of the AfriCOMET or AfriCOMET-QE systems. As showcased in Table~\ref{tab:wmt_afr_eval_result}, in the MT evaluation task, AfriCOMET-STL surpasses the cutting-edge COMET22 system across all three correlation coefficients. Meanwhile, AfriCOMET-MTL shows a slight edge over COMET22 in the Pearson correlation coefficient. For QE, both AfriCOMET-QE-STL and AfriCOMET-QE-MTL significantly outperform the state-of-the-art CometKiwi system. These comparisons are fair since all systems are trained using the ``WMT Others'' dataset. This evaluation further validates the efficacy of our benchmark systems from an additional perspective.

\subsection{Evaluation on the WMT 2022 English-Yoruba QE test set}
\label{subsec:appendix c}
The WMT organizers recently released an English-Yoruba DA dataset, serving as the zero-shot test set in the WMT 2022 Quality Estimation Shared Task. This dataset consists of 1010 DA annotations, prepared using DA guidelines different from ours, as outlined by~\citet{fomicheva-etal-2021-eval4nlp}. The source sentences are sampled from Wikipedia, covering seven topics, and translated into Yoruba using Google Translate, as reported in~\citet{zerva2022findings}. We evaluate CometKiwi and our benchmark AfriCOMET-QE systems on this dataset. The results, shown in Table~\ref{tab:wmt22-qe-eng-yor-result}, demonstrate that our AfriCOMET-QE systems outperform CometKiwi significantly on this English-Yoruba dataset, underscoring the efficacy of our QE approaches even with differently guided DA annotations.

\subsection{Generalization Evaluation}
\label{subsec:generalization}
Given that our benchmark systems employ the African language-enhanced pre-trained model, AfroXLM-R-L, assessing their generalization capabilities on non-African datasets is crucial. The development of both COMET22 and CometKiwi systems involves using the English-German, English-Russian, and Chinese-English MQM datasets from the WMT 2022 News Domain Translation Shared Task as validation sets, featuring 8959, 8432, and 9750 MQM annotations, respectively. Therefore, testing our benchmark systems on these three datasets is practical to evaluate their generalization in non-African cases. We present the results of correlation coefficients in Table~\ref{tab:generalization_result}. In MT evaluation and QE tasks, AfriCOMET and AfriCOMET-QE exhibit only a slight performance drop compared to COMET22 and CometKiwi systems, respectively, which might be due to the adaptation feasibility of the AfroXLM-R-L pre-trained encoder. This evaluation highlights the sustained generalization capabilities of our benchmark systems.

\input{tables/wmt_afr_eval}
\input{tables/wmt22-qe-eng-yor}
\input{tables/generalization}

%% file: tables/lang_family.tex
\begin{table*}[!t]
\small
\begin{center}
\centering
\resizebox{1.0\columnwidth}{!}{%
\begin{tabular}{l|l}
\toprule
\textbf{Language} & \textbf{Family Group} \\
\midrule
Darija	&	 Afro-Asiatic/Semitic	\\
Egyptian Arabic	&	 Afro-Asiatic/Semitic	\\
English	&	 Indo-European/Germanic/Anglo-Frisian	\\
French	&	 Indo-European/Italic/Romance	\\
Hausa	&	 Afro-Asiatic/Chadic	\\
Igbo	&	 Atlantic-Congo/Volta-Niger	\\
Kikuyu	&	 Atlantic-Congo/Bantu/North-East Bantu	\\
Luo	&	 Nilo-Saharan/Nilotic	\\
Somali	&	 Afro-Asiatic/Cushitic	\\
Swahili	&	 Atlantic-Congo/Bantu/North-East Bantu	\\
Twi	&	 Atlantic-Congo/Kwa	\\
Xhosa	&	 Atlantic-Congo/Bantu/Southern Bantu/Nguni	\\
Yoruba	&	 Atlantic-Congo/Volta-Niger	\\
\bottomrule
\end{tabular}
}
\end{center}
\caption{Language family groups that our targeted African languages belong to, according to Wikipedia (\url{https://en.wikipedia.org/wiki/Language_family}).}
\label{tab:language_family}
\end{table*}

%% file: tables/overall_count.tex
\begin{table*}[!t]
\begin{center}
\centering
\resizebox{1.0\columnwidth}{!}{%
\begin{tabular}{l|c|c|cc}
\toprule
LP & original \# & qualified \# & dev \# & devtest \# \\
\midrule
ary-fra	& 520&	394	&	207	&	187	\\
eng-arz	& 520&	518	&	268	&	250	\\
eng-fra	& 520&	515	&	265	&	250	\\
eng-hau	& 520&	490	&	250	&	240	\\
eng-ibo	& 520&	240	&	120	&	120	\\
eng-kik	& 520&	410	&	208	&	202	\\
eng-luo	& 520&	499	&	257	&	242	\\
eng-som	& 520&	434	&	208	&	226	\\
eng-swh	& 520&	352	&	195	&	157	\\
eng-twi	& 520&	516	&	269	&	247	\\
eng-xho	& 520&	494	&	251	&	243	\\
eng-yor	& 520&	484	&	245	&	239	\\
eng-yor (it)	& 250&	217	&	-	&	217	\\
eng-yor (movie)	& 270&	219	&	-	&	219	\\
eng-yor (news)	& 270&	237	&	-	&	237	\\
eng-yor (ted)	& 250&	224	&	-	&	224	\\
yor-eng	& 520&	439	&	227	&	212	\\
\bottomrule
\end{tabular}
}
\end{center}
\caption{Counts of qualified~\textbf{adequacy} annotations for each language pair in dev and devtest sets, with English-Yoruba exclusively as devtest in domain-Specific datasets.}
\label{tab:overall_count}
\end{table*}

%% file: tables/overall_count_fluency.tex
\begin{table*}[!t]
\begin{center}
\centering
\resizebox{1.0\columnwidth}{!}{%
\begin{tabular}{l|c|c|cc}
\toprule
LP & original \# & qualified \# & dev \# & devtest \# \\
\midrule
ary-fra	&	520	&	459	&	239	&	220	\\
eng-arz	&	520	&	518	&	268	&	250	\\
eng-fra	&	520	&	459	&	244	&	215	\\
eng-hau	&	520	&	482	&	234	&	248	\\
eng-ibo	&	520	&	409	&	178	&	231	\\
eng-kik	&	-	&	-	&	-	&	-	\\
eng-luo	&	-	&	-	&	-	&	-	\\
eng-som	&	520	&	450	&	224	&	226	\\
eng-swh	&	520	&	376	&	177	&	199	\\
eng-twi	&	520	&	518	&	269	&	249	\\
eng-xho	&	520	&	497	&	250	&	247	\\
eng-yor	&	520	&	495	&	261	&	234	\\
eng-yor (it)	&	250	&	237	&	-	&	237	\\
eng-yor (movie)	&	270	&	262	&	-	&	262	\\
eng-yor (news)	&	270	&	258	&	-	&	258	\\
eng-yor (ted)	&	250	&	243	&	-	&	243	\\
yor-eng	&	520	&	500	&	258	&	242	\\
\bottomrule
\end{tabular}
}
\end{center}
\caption{Counts of qualified~\textbf{fluency} annotations for each language pair in dev and devtest sets, with English-Yoruba exclusively as devtest in domain-specific datasets.}
\label{tab:overall_count_fluency}
\end{table*}

%% file: tables/wmt_previous.tex
\begin{table*}[!t]
\begin{center}
\centering
\resizebox{1.0\columnwidth}{!}{%
\begin{tabular}{l|cccc}
\toprule
LP	&	Annotation Count	&	Median	&	Mean	&	Std	\\
\midrule
ces-eng	&	27847	&	75.00	&	69.12	&	25.18	\\
deu-ces	&	13804	&	56.00	&	53.35	&	32.97	\\
deu-eng	&	99183	&	81.00	&	73.00	&	27.06	\\
deu-fra	&	6691	&	78.00	&	71.04	&	27.44	\\
eng-ces	&	60937	&	69.00	&	62.48	&	29.09	\\
eng-deu	&	121420	&	90.00	&	80.79	&	23.2	\\
eng-est	&	13376	&	51.00	&	51.82	&	29.83	\\
eng-fin	&	34335	&	53.00	&	53.04	&	30.3	\\
eng-guj	&	6924	&	48.50	&	49.70	&	28.16	\\
eng-jpn	&	9578	&	72.67	&	68.31	&	20.45	\\
eng-kaz	&	8219	&	57.50	&	54.16	&	28.86	\\
eng-lit	&	8959	&	60.00	&	57.40	&	29.77	\\
eng-lvs	&	5810	&	40.00	&	43.09	&	29.36	\\
eng-mar	&	26000	&	71.75	&	70.08	&	10.15	\\
eng-pol	&	10572	&	74.00	&	69.57	&	22.36	\\
eng-rus	&	62749	&	75.00	&	67.98	&	27.26	\\
eng-tam	&	7890	&	74.00	&	70.06	&	19.14	\\
eng-tur	&	5171	&	50.00	&	48.10	&	33.92	\\
eng-zho	&	90805	&	77.00	&	73.65	&	20.27	\\
est-eng	&	29496	&	70.00	&	63.48	&	28.85	\\
fin-eng	&	46145	&	75.00	&	66.29	&	29.17	\\
fra-deu	&	3999	&	83.00	&	76.13	&	23.86	\\
guj-eng	&	9063	&	58.00	&	55.70	&	29.61	\\
jpn-eng	&	8939	&	76.00	&	70.72	&	24.8	\\
kaz-eng	&	6789	&	72.00	&	64.72	&	28.09	\\
khm-eng	&	4722	&	69.00	&	61.60	&	28.01	\\
lit-eng	&	10315	&	77.00	&	70.23	&	25.31	\\
npi-eng	&	9000	&	33.67	&	37.92	&	19.51	\\
pol-eng	&	11816	&	80.12	&	76.14	&	21.62	\\
pbt-eng	&	4611	&	70.00	&	64.14	&	25.61	\\
ron-eng	&	9000	&	76.33	&	68.76	&	27.31	\\
rus-eng	&	79280	&	84.00	&	75.38	&	25.24	\\
sin-eng	&	9000	&	50.00	&	50.45	&	28.33	\\
tam-eng	&	7577	&	72.00	&	65.45	&	26.68	\\
tur-eng	&	30186	&	71.00	&	63.51	&	29.17	\\
zho-eng	&	126947	&	79.00	&	73.37	&	24.67	\\
\midrule
Total.	&	1027155	&		&		&		\\
\bottomrule
\end{tabular}
}
\end{center}
\caption{Statistical summary of~\textbf{WMT Others} across language pairs: annotation counts, and the median, mean, and standard deviation of the DA scores. Language codes correspond to those specified in FLORES-200~\citep{goyal2022flores}.}
\label{tab:wmt_previous}
\end{table*}

%% file: tables/wmt_african.tex
\begin{table*}[!t]
\begin{center}
\centering
\resizebox{1.0\columnwidth}{!}{%
\begin{tabular}{l|cccc}
\toprule
LP	&	Annotation Count	&	Median	&	Mean	&	Std	\\
\midrule
afr-eng	&	778	&	78.0	&	64.14	&	32.1	\\
afr-ssw	&	594	&	68.0	&	55.32	&	29.76	\\
amh-eng	&	594	&	72.5	&	60.32	&	33.4	\\
eng-afr	&	593	&	63.0	&	62.23	&	30.74	\\
eng-amh	&	594	&	55.0	&	48.37	&	27.87	\\
eng-hau	&	592	&	69.0	&	58.58	&	38	\\
eng-ibo	&	593	&	71.0	&	53.59	&	42.6	\\
eng-kin	&	594	&	57.5	&	53.60	&	38.32	\\
eng-lug	&	594	&	60.0	&	51.05	&	38.02	\\
eng-nya	&	594	&	81.0	&	60.44	&	39.92	\\
eng-orm	&	594	&	43.5	&	43.80	&	34.17	\\
eng-sna	&	593	&	92.0	&	75.79	&	36.3	\\
eng-ssw	&	594	&	58.0	&	50.87	&	33.69	\\
eng-swh	&	591	&	85.0	&	71.13	&	32.83	\\
eng-tsn	&	792	&	80.0	&	64.48	&	35.6	\\
eng-xho	&	594	&	87.5	&	61.87	&	37.56	\\
eng-yor	&	594	&	71.0	&	57.79	&	35.29	\\
eng-zul	&	792	&	84.0	&	66.19	&	38.45	\\
fra-lin	&	594	&	89.0	&	70.83	&	36.68	\\
fra-swh	&	592	&	65.0	&	56.70	&	30.04	\\
hau-eng	&	789	&	83.0	&	69.94	&	32.36	\\
hau-ibo	&	594	&	48.0	&	46.74	&	38.42	\\
ibo-eng	&	790	&	82.0	&	61.38	&	38.45	\\
ibo-hau	&	593	&	69.0	&	51.78	&	37.19	\\
ibo-yor	&	594	&	52.0	&	45.48	&	36.52	\\
kin-eng	&	590	&	84.0	&	65.21	&	38.05	\\
lin-fra	&	592	&	86.5	&	69.66	&	36.5	\\
lug-eng	&	792	&	42.0	&	45.95	&	35.54	\\
nya-eng	&	594	&	70.0	&	58.20	&	34.64	\\
orm-eng	&	594	&	23.0	&	40.93	&	39.88	\\
sna-eng	&	784	&	91.0	&	78.65	&	31.58	\\
som-eng	&	594	&	70.0	&	58.17	&	34.95	\\
ssw-eng	&	791	&	80.0	&	62.11	&	40.01	\\
ssw-tsn	&	594	&	75.5	&	66.37	&	28.07	\\
swh-eng	&	779	&	86.0	&	71.26	&	33.02	\\
swh-fra	&	591	&	83.0	&	68.68	&	31.65	\\
swh-lug	&	594	&	14.0	&	30.40	&	33.41	\\
tsn-eng	&	791	&	63.0	&	54.25	&	35.24	\\
tsn-tso	&	594	&	70.5	&	63.66	&	29.68	\\
tso-eng	&	787	&	70.0	&	59.34	&	36.18	\\
xho-eng	&	789	&	85.0	&	71.72	&	31.83	\\
xho-zul	&	594	&	68.0	&	49.45	&	36.56	\\
yor-eng	&	792	&	63.0	&	57.45	&	33.69	\\
yor-ibo	&	594	&	80.0	&	67.69	&	33.09	\\
zul-eng	&	788	&	90.0	&	68.47	&	38.54	\\
zul-sna	&	593	&	82.0	&	64.89	&	42.39	\\
\midrule
Total.	&	30021	&		&		&		\\
\bottomrule
\end{tabular}
}
\end{center}
\caption{Statistical summary of~\textbf{WMT African} across language pairs: annotation counts, and the median, mean, and standard deviation of DA scores. Language codes correspond to those specified in FLORES-200~\citep{goyal2022flores}.}
\label{tab:wmt_african}
\end{table*}

%% file: tables/language_coverage.tex
\begin{table*}[!t]
\small
\begin{center}
\centering
\resizebox{2.0\columnwidth}{!}{%
\begin{tabular}{l|c|c}
\toprule
\textbf{Pre-trained Encoder} & \textbf{Languages Covered} & \textbf{Languages Uncovered} \\
\midrule
XLM-R-L	& 
English, French, Arabic, Hausa, Somali, Swahili, Xhosa & 
Igbo, Luo, Kikuyu, Twi, Yoruba \\
InfoXLM-R-L	&
English, French, Arabic, Hausa, Somali, Swahili, Xhosa	&
Igbo, Luo, Kikuyu, Twi, Yoruba \\
AfroXLM-R-L	&
English, French, Arabic, Hausa, Igbo, Somali, Swahili, Xhosa, Yoruba & 
Luo, Kikuyu, Twi \\
\bottomrule
\end{tabular}
}
\end{center}
\caption{Overview of language coverage for XLM-Roberta-Large (XLM-R-L)~\citep{conneau2019unsupervised}, InfoXLM-Large (InfoXLM-L)~\citep{chi2020infoxlm}, and AfroXLM-Roberta-Large (AfroXLM-R-L)~\citep{alabi-etal-2022-adapting} as utilized in this study.}
\label{tab:language coverage}
\end{table*}

%% file: tables/pearson_kendall_main.tex
\begin{table*}[!t]
\centering
\scalebox{0.56}{
\begin{tabular}{l|cc|c|c|c|ccc|c}
\toprule
& \multicolumn{2}{c|}{\textbf{N-gram Matching}} 
& \multicolumn{1}{c|}{\textbf{Embedding-based}} 
& \multicolumn{1}{c|}{\textbf{LLM Prompting}}
& \multicolumn{5}{c}{\textbf{Learned COMET Metric}} \\
\cmidrule{2-10}
& \multicolumn{1}{c}{\multirow{2}{*}{\textbf{SacreBLEU}}} 
& \multicolumn{1}{c|}{\multirow{2}{*}{\textbf{chrF++}}}
& \multicolumn{1}{c|}{\multirow{2}{*}{\textbf{BERTScore}}}
& \multicolumn{1}{c|}{\multirow{2}{*}{\textbf{GPT4}}} 
& \textbf{Baseline} 
& \multicolumn{3}{c|}{\textbf{Single Task} (Ours)} 
& \textbf{Multi Task} (Ours) \\
\cmidrule{6-10}
LP & & & & & COMET22 & XLM-R-L & InfoXLM-L & AfroXLM-R-L \ding{72} & AfroXLM-R-L \ding{72} \\
\midrule
ary-fra	&	0.307 / 0.234	&	0.402 / 0.233	&	0.414 / 0.242	&	\textbf{0.693} / \textbf{0.467}	&	0.584 / 0.379	&	0.634 / 0.397	&	0.631 / \textbf{0.406}	&	0.595 / \textbf{0.406}	&	\textbf{0.685} / \textbf{0.447}	\\
eng-arz	&	0.241 / 0.222	&	0.290 / 0.214	&	0.314 / 0.234	&	0.454 / \textbf{0.379}	&	0.528 / 0.347	&	0.533 / 0.339	&	0.498 / 0.337	&	0.526 / 0.371	&	\textbf{0.602} / \textbf{0.423}	\\
eng-fra	&	0.268 / 0.171	&	0.339 / 0.193	&	0.358 / 0.195	&	\textbf{0.495} / \textbf{0.385}	&	\textbf{0.475} / \textbf{0.344}	&	\textbf{0.469} / \textbf{0.359}	&	\textbf{0.443} / 0.324	&	\textbf{0.515} / \textbf{0.351}	&	\textbf{0.522} / \textbf{0.372}	\\
eng-hau	&	0.248 / 0.137	&	0.445 / 0.206	&	0.576 / 0.283	&	\textbf{0.664} / 0.278	&	0.589 / 0.302	&	0.503 / 0.286	&	0.473 / 0.229	&	\textbf{0.682} / 0.365	&	\textbf{0.696} / \textbf{0.445}	\\
eng-ibo	&	0.304 / 0.235	&	0.475 / 0.294	&	0.365 / 0.292	&	0.466 / 0.194	&	0.323 / 0.259	&	0.386 / 0.288	&	0.312 / 0.260	&	\textbf{0.551} / \textbf{0.435}	&	\textbf{0.649} / \textbf{0.445}	\\
eng-kik	&	0.256 / 0.187	&	0.406 / \textbf{0.202}	&	\textbf{0.498} / 0.188	&	0.448 / \textbf{0.196}	&	0.434 / 0.139	&	0.464 / 0.186	&	0.393 / 0.169	&	\textbf{0.582} / \textbf{0.270}	&	\textbf{0.523} / \textbf{0.276}	\\
eng-luo	&	0.182 / 0.122	&	0.320 / 0.187	&	\textbf{0.429} / \textbf{0.250}	&	0.222 / 0.183	&	0.203 / 0.039	&	0.258 / 0.136	&	0.354 / 0.166	&	\textbf{0.427} / \textbf{0.191}	&	\textbf{0.433} / \textbf{0.251}	\\
eng-som	&	0.170 / 0.108	&	0.317 / 0.196	&	0.298 / 0.240	&	\textbf{0.485} / 0.205	&	\textbf{0.526} / 0.338	&	\textbf{0.503} / 0.334	&	\textbf{0.465} / 0.297	&	\textbf{0.470} / \textbf{0.398}	&	\textbf{0.391} / \textbf{0.389}	\\
eng-swh	&	0.459 / 0.334	&	0.648 / 0.408	&	\textbf{0.773} / 0.516	&	\textbf{0.768} / \textbf{0.604}	&	\textbf{0.779} / \textbf{0.560}	&	\textbf{0.771} / \textbf{0.567}	&	\textbf{0.775} / \textbf{0.546}	&	0.729 / 0.508	&	\textbf{0.754} / \textbf{0.552}	\\
eng-twi	&	0.185 / \textbf{0.137}	&	0.223 / \textbf{0.120}	&	0.292 / 0.074	&	\textbf{0.456} / \textbf{0.096}	&	\textbf{0.378} / 0.064	&	\textbf{0.341} / 0.070	&	0.274 / \textbf{0.078}	&	\textbf{0.396} / \textbf{0.104}	&	0.295 / 0.071	\\
eng-xho	&	0.124 / 0.072	&	0.246 / \textbf{0.128}	&	0.306 / \textbf{0.132}	&	\textbf{0.433} / \textbf{0.117}	&	0.234 / 0.055	&	0.202 / 0.054	&	0.278 / 0.046	&	\textbf{0.473} / \textbf{0.150}	&	\textbf{0.465} / \textbf{0.115}	\\
eng-yor	&	0.236 / 0.144	&	0.355 / 0.143	&	0.462 / 0.176	&	\textbf{0.674} / \textbf{0.334}	&	0.367 / 0.103	&	0.329 / 0.131	&	0.353 / 0.129	&	0.463 / 0.201	&	\textbf{0.694} / 0.256	\\
eng-yor (it)	&	0.219 / 0.206	&	0.411 / 0.244	&	\textbf{0.659} / \textbf{0.297}	&	\textbf{0.626} / \textbf{0.327}	&	\textbf{0.660} / 0.233	&	0.558 / 0.177	&	0.614 / 0.184	&	0.590 / 0.183	&	\textbf{0.659} / \textbf{0.298}	\\
eng-yor (movie)	&	0.224 / 0.166	&	0.288 / 0.152	&	0.430 / 0.213	&	\textbf{0.630} / \textbf{0.403}	&	0.486 / 0.237	&	0.429 / 0.240	&	0.503 / 0.256	&	0.464 / 0.261	&	0.501 / 0.268	\\
eng-yor (news)	&	0.207 / 0.081	&	0.294 / 0.086	&	0.366 / 0.075	&	\textbf{0.521} / \textbf{0.144}	&	0.395 / \textbf{0.118}	&	0.373 / \textbf{0.137}	&	0.392 / \textbf{0.090}	&	\textbf{0.508} / \textbf{0.136}	&	\textbf{0.501} / \textbf{0.147}	\\
eng-yor (ted)	&	0.037 / 0.019	&	0.100 / 0.002	&	0.284 / 0.062	&	\textbf{0.451} / \textbf{0.176}	&	0.351 / 0.083	&	0.377 / 0.122	&	0.449 / \textbf{0.185}	&	\textbf{0.539} / \textbf{0.224}	&	0.408 / \textbf{0.207}	\\
yor-eng	&	0.257 / 0.208	&	0.389 / 0.281	&	0.425 / 0.308	&	0.464 / \textbf{0.338}	&	\textbf{0.508} / \textbf{0.354}	&	0.452 / 0.323	&	\textbf{0.486} / 0.335	&	\textbf{0.512} / \textbf{0.345}	&	\textbf{0.544} / \textbf{0.382}	\\
\midrule
Avg.	&	0.231 / 0.164	&	0.350 / 0.193	&	0.426 / 0.222	&	0.526 / 0.284	&	0.460 / 0.233	&	0.446 / 0.244	&	0.453 / 0.237	&	\textbf{0.531} / \textbf{0.288}	&	\textbf{0.548} / \textbf{0.314}	\\
\bottomrule
\end{tabular}
}
\caption{Sentence-level Pearson and Kendall-rank correlation coefficients for MT evaluation models.
For each LP, values in~\textbf{bold} represent the highest ranking obtained from the Perm-Input hypothesis test~\citep{deutsch2021statistical}. Comprehensive results of this test are detailed in Table~\ref{tab:pearson_main_rank} and~\ref{tab:kendall_main_rank}. Averaged Pearson and Kendall-rank correlations across LPs are presented in the last row.}
\label{tab:pearson_kendall_main_result}
\end{table*}

%% file: tables/spearman_main_rank.tex
\begin{table*}[!t]
\begin{center}
\centering
\resizebox{2.0\columnwidth}{!}{%
\begin{tabular}{l|cc|c|c|c|ccc|c}
\toprule
& \multicolumn{2}{c|}{\textbf{N-gram Matching}} 
& \multicolumn{1}{c|}{\textbf{Embedding-based}} 
& \multicolumn{1}{c|}{\textbf{LLM Prompting}}
& \multicolumn{5}{c}{\textbf{Learned COMET Metric}} \\
\cmidrule{2-10}
& \multicolumn{1}{c}{\multirow{2}{*}{SacreBLEU}} 
& \multicolumn{1}{c|}{\multirow{2}{*}{chrF++}}
& \multicolumn{1}{c|}{\multirow{2}{*}{BERTScore}}
& \multicolumn{1}{c|}{\multirow{2}{*}{GPT-4}} 
& \textbf{Baseline} 
& \multicolumn{3}{c|}{\textbf{Single Task} (Ours)} 
& \textbf{Multi Task} (Ours) \\
\cmidrule{6-10}
LP & & & & & COMET22 & XLM-R-L & InfoXLM-L & AfroXLM-R-L \ding{72} & AfroXLM-R-L \ding{72} \\
\midrule
ary-fra	&	3	&	3	&	3	&	1	&	2	&	2	&	1	&	1	&	1	\\
en-arz	&	4	&	4	&	4	&	2	&	2	&	3	&	3	&	2	&	1	\\
en-fra	&	3	&	3	&	3	&	1	&	1	&	1	&	2	&	1	&	1	\\
en-hau	&	5	&	4	&	3	&	3	&	2	&	3	&	3	&	2	&	1	\\
en-ibo	&	2	&	2	&	2	&	3	&	2	&	2	&	2	&	1	&	1	\\
en-kik	&	2	&	1	&	2	&	2	&	3	&	2	&	2	&	1	&	1	\\
en-luo	&	3	&	2	&	1	&	2	&	3	&	2	&	2	&	1	&	1	\\
en-som	&	5	&	4	&	3	&	3	&	2	&	2	&	2	&	1	&	1	\\
en-swh	&	4	&	3	&	2	&	1	&	1	&	1	&	1	&	2	&	1	\\
en-twi	&	1	&	1	&	2	&	1	&	2	&	2	&	1	&	1	&	2	\\
en-xho	&	2	&	1	&	1	&	1	&	2	&	2	&	2	&	1	&	1	\\
en-yor 	&	3	&	3	&	2	&	1	&	4	&	3	&	3	&	2	&	2	\\
en-yor (it)	&	2	&	2	&	1	&	1	&	2	&	3	&	2	&	2	&	1	\\
en-yor (movie)	&	3	&	3	&	3	&	1	&	2	&	2	&	2	&	2	&	2	\\
en-yor (news)	&	2	&	2	&	2	&	1	&	1	&	1	&	1	&	1	&	1	\\
en-yor (ted)	&	3	&	3	&	2	&	1	&	2	&	2	&	1	&	1	&	1	\\
yor-eng	&	3	&	2	&	2	&	1	&	1	&	2	&	1	&	1	&	1	\\
\midrule
Avg.	&	2.94	&	2.53	&	2.24	&	1.53	&	2.00	&	2.06	&	1.82	&	\textbf{1.35}	&	\textbf{1.18}	\\
\bottomrule
\end{tabular}
}
\end{center}
\caption{Detailed rankings from the Perm-Input hypothesis test~\citep{deutsch2021statistical} of Spearman-rank correlation coefficients corresponding to Table~\ref{tab:spearman_main_result}. The averaged ranks are presented in the last row.}
\label{tab:spearman_main_rank}
\end{table*}

\begin{table*}[!t]
\begin{center}
\centering
\resizebox{2.0\columnwidth}{!}{%
\begin{tabular}{l|cc|c|c|c|ccc|c}
\toprule
& \multicolumn{2}{c|}{\textbf{N-gram Matching}} 
& \multicolumn{1}{c|}{\textbf{Embedding-based}} 
& \multicolumn{1}{c|}{\textbf{LLM Prompting}}
& \multicolumn{5}{c}{\textbf{Learned COMET Metric}} \\
\cmidrule{2-10}
& \multicolumn{1}{c}{\multirow{2}{*}{SacreBLEU}} 
& \multicolumn{1}{c|}{\multirow{2}{*}{chrf++}}
& \multicolumn{1}{c|}{\multirow{2}{*}{BERTScore}}
& \multicolumn{1}{c|}{\multirow{2}{*}{GPT-4}} 
& \textbf{Baseline} 
& \multicolumn{3}{c|}{\textbf{Single Task} (Ours)} 
& \textbf{Multi Task} (Ours) \\
\cmidrule{6-10}
LP & & & & & COMET22 & XLM-R-L & InfoXLM-L & AfroXLM-R-L \ding{72} & AfroXLM-R-L \ding{72} \\
\midrule
ary-fra	&	4	&	3	&	3	&	1	&	2	&	2	&	2	&	2	&	1	\\
eng-arz	&	3	&	3	&	3	&	2	&	2	&	2	&	2	&	2	&	1	\\
eng-fra	&	3	&	2	&	2	&	1	&	1	&	1	&	1	&	1	&	1	\\
eng-hau	&	4	&	3	&	2	&	1	&	2	&	3	&	3	&	1	&	1	\\
eng-ibo	&	3	&	2	&	3	&	2	&	3	&	2	&	3	&	1	&	1	\\
eng-kik	&	3	&	2	&	1	&	2	&	2	&	2	&	3	&	1	&	1	\\
eng-luo	&	4	&	2	&	1	&	3	&	4	&	3	&	2	&	1	&	1	\\
eng-som	&	3	&	2	&	2	&	1	&	1	&	1	&	1	&	1	&	1	\\
eng-swh	&	3	&	2	&	1	&	1	&	1	&	1	&	1	&	2	&	1	\\
eng-twi	&	2	&	2	&	2	&	1	&	1	&	1	&	2	&	1	&	2	\\
eng-xho	&	3	&	2	&	2	&	1	&	2	&	3	&	2	&	1	&	1	\\
eng-yor 	&	4	&	3	&	2	&	1	&	3	&	4	&	3	&	2	&	1	\\
eng-yor (it)	&	5	&	4	&	1	&	1	&	1	&	3	&	2	&	2	&	1	\\
eng-yor (movie)	&	4	&	4	&	3	&	1	&	2	&	3	&	2	&	2	&	2	\\
eng-yor (news)	&	3	&	2	&	2	&	1	&	2	&	2	&	2	&	1	&	1	\\
eng-yor (ted)	&	5	&	5	&	4	&	1	&	3	&	3	&	2	&	1	&	2	\\
yor-eng	&	3	&	2	&	2	&	2	&	1	&	2	&	1	&	1	&	1	\\
\midrule
Avg.	&	3.47	&	2.65	&	2.12	&	\textbf{1.35}	&	1.94	&	2.24	&	2.00	&	\textbf{1.35}	&	\textbf{1.18}	\\
\bottomrule
\end{tabular}
}
\end{center}
\caption{Detailed rankings from the Perm-Input hypothesis test~\citep{deutsch2021statistical} of Pearson correlation coefficients corresponding to Table~\ref{tab:pearson_kendall_main_result}. The averaged ranks are presented in the last row.}
\label{tab:pearson_main_rank}
\end{table*}

\begin{table*}[!t]
\begin{center}
\centering
\resizebox{2.0\columnwidth}{!}{%
\begin{tabular}{l|cc|c|c|c|ccc|c}
\toprule
& \multicolumn{2}{c|}{\textbf{N-gram Matching}} 
& \multicolumn{1}{c|}{\textbf{Embedding-based}} 
& \multicolumn{1}{c|}{\textbf{LLM Prompting}}
& \multicolumn{5}{c}{\textbf{Learned COMET Metric}} \\
\cmidrule{2-10}
& \multicolumn{1}{c}{\multirow{2}{*}{SacreBLEU}} 
& \multicolumn{1}{c|}{\multirow{2}{*}{chrf++}}
& \multicolumn{1}{c|}{\multirow{2}{*}{BERTScore}}
& \multicolumn{1}{c|}{\multirow{2}{*}{GPT-4}} 
& \textbf{Baseline} 
& \multicolumn{3}{c|}{\textbf{Single Task} (Ours)} 
& \textbf{Multi Task} (Ours) \\
\cmidrule{6-10}
LP & & & & & COMET22 & XLM-R-L & InfoXLM-L & AfroXLM-R-L \ding{72} & AfroXLM-R-L \ding{72} \\
\midrule
ary-fra	&	3	&	3	&	3	&	1	&	2	&	2	&	1	&	1	&	1	\\
eng-arz	&	4	&	4	&	4	&	1	&	2	&	2	&	3	&	2	&	1	\\
eng-fra	&	3	&	3	&	3	&	1	&	1	&	1	&	2	&	1	&	1	\\
eng-hau	&	5	&	4	&	3	&	3	&	2	&	3	&	4	&	2	&	1	\\
eng-ibo	&	2	&	2	&	2	&	3	&	2	&	2	&	2	&	1	&	1	\\
eng-kik	&	2	&	1	&	2	&	1	&	2	&	2	&	2	&	1	&	1	\\
eng-luo	&	3	&	2	&	1	&	2	&	3	&	2	&	2	&	1	&	1	\\
eng-som	&	4	&	3	&	3	&	3	&	2	&	2	&	2	&	1	&	1	\\
eng-swh	&	4	&	3	&	2	&	1	&	1	&	1	&	1	&	2	&	1	\\
eng-twi	&	1	&	1	&	2	&	1	&	2	&	2	&	1	&	1	&	2	\\
eng-xho	&	2	&	1	&	1	&	1	&	2	&	2	&	2	&	1	&	1	\\
eng-yor &	3	&	3	&	3	&	1	&	4	&	3	&	3	&	2	&	2	\\
eng-yor (it)	&	2	&	2	&	1	&	1	&	2	&	3	&	2	&	2	&	1	\\
eng-yor (movie)	&	3	&	3	&	2	&	1	&	2	&	2	&	2	&	2	&	2	\\
eng-yor (news)	&	2	&	2	&	2	&	1	&	1	&	1	&	1	&	1	&	1	\\
eng-yor (ted)	&	3	&	3	&	2	&	1	&	2	&	2	&	1	&	1	&	1	\\
yor-eng	&	3	&	2	&	2	&	1	&	1	&	2	&	2	&	1	&	1	\\
\midrule
Avg.	&	2.88	&	2.47	&	2.24	&	1.41	&	1.94	&	2.00	&	1.94	&	\textbf{1.35}	&	\textbf{1.18}	\\

\bottomrule
\end{tabular}
}
\end{center}
\caption{Detailed rankings from the Perm-Input hypothesis test~\citep{deutsch2021statistical} of Kendall-rank correlation coefficients corresponding to Table~\ref{tab:pearson_kendall_main_result}. The averaged ranks are presented in the last row.}
\label{tab:kendall_main_rank}
\end{table*}

%% file: tables/afr_xlmr_result.tex
\begin{table*}[t]
\begin{center}
\small
\centering
\resizebox{2.0\columnwidth}{!}{%
\begin{tabular}{l|ccc|ccc|ccc}
\toprule
& \multicolumn{9}{c}{Training Data Settings} \\
\cmidrule{2-10}
& \multicolumn{3}{c|}{WMT African} & \multicolumn{3}{c|}{WMT Others} & \multicolumn{3}{c}{WMT Combined}  \\
\midrule
LP & Pearson & Spearman & Kendall & Pearson & Spearman & Kendall & Pearson & Spearman & Kendall \\
\midrule
ary-fra	&	0.307	&	0.287	&	0.201	&	0.595	&	0.567	&	0.406	&	0.567	&	0.547	&	0.388	\\
eng-arz	&	0.215	&	0.270	&	0.177	&	0.526	&	0.532	&	0.371	&	0.517	&	0.506	&	0.351	\\
eng-fra	&	0.380	&	0.276	&	0.190	&	0.515	&	0.495	&	0.351	&	0.545	&	0.501	&	0.355	\\
eng-hau	&	0.676	&	0.354	&	0.240	&	0.682	&	0.515	&	0.365	&	0.764	&	0.489	&	0.342	\\
eng-ibo	&	0.357	&	0.406	&	0.290	&	0.551	&	0.592	&	0.435	&	0.452	&	0.562	&	0.417	\\
eng-kik	&	0.618	&	0.256	&	0.172	&	0.582	&	0.389	&	0.270	&	0.654	&	0.368	&	0.254	\\
eng-luo	&	0.416	&	0.255	&	0.181	&	0.427	&	0.283	&	0.191	&	0.404	&	0.275	&	0.187	\\
eng-som	&	0.479	&	0.388	&	0.271	&	0.470	&	0.554	&	0.398	&	0.590	&	0.546	&	0.390	\\
eng-swh	&	0.642	&	0.533	&	0.373	&	0.729	&	0.688	&	0.508	&	0.735	&	0.692	&	0.515	\\
eng-twi	&	0.436	&	0.124	&	0.082	&	0.396	&	0.157	&	0.104	&	0.484	&	0.203	&	0.139	\\
eng-xho	&	0.519	&	0.092	&	0.072	&	0.473	&	0.191	&	0.150	&	0.573	&	0.200	&	0.155	\\
eng-yor  &	0.597	&	0.127	&	0.083	&	0.463	&	0.287	&	0.201	&	0.668	&	0.285	&	0.202	\\
eng-yor (it)	&	0.712	&	0.251	&	0.172	&	0.590	&	0.266	&	0.183	&	0.797	&	0.247	&	0.172	\\
eng-yor (movie)	&	0.550	&	0.274	&	0.188	&	0.464	&	0.372	&	0.261	&	0.613	&	0.349	&	0.242	\\
eng-yor (news)	&	0.468	&	0.066	&	0.045	&	0.508	&	0.200	&	0.136	&	0.614	&	0.204	&	0.141	\\
eng-yor (ted)	&	0.404	&	0.084	&	0.058	&	0.539	&	0.324	&	0.224	&	0.608	&	0.220	&	0.151	\\
yor-eng	&	0.406	&	0.386	&	0.256	&	0.512	&	0.490	&	0.345	&	0.511	&	0.495	&	0.346	\\

\midrule

Avg. &	0.481 	&	0.261 	&	0.179 	&	0.531 	&	\textbf{0.406} 	&	\textbf{0.288} 	&	\textbf{0.594} 	&	0.393 	&	0.279 	\\

\bottomrule
\end{tabular}
}
\end{center}
\caption{Correlation coefficients (Pearson, Spearman-rank, Kendall-rank) for MT evaluation models trained with \textbf{single-task learning} based on AfroXLM-R-L with varied training data settings. Comprehensive results of the Perm-Input hypothesis test~\citep{deutsch2021statistical} are detailed in Table~\ref{tab:afr_xlmr_stl_rank}. The averaged correlation coefficients are presented in the last row.}
\label{tab:afr_xlmr_result}
\end{table*}

%% file: tables/afr_xlmr_stl_rank.tex
\begin{table*}[t]
\begin{center}
\small
\centering
\resizebox{2.0\columnwidth}{!}{%
\begin{tabular}{l|ccc|ccc|ccc}
\toprule
& \multicolumn{9}{c}{Training Data Settings} \\
\cmidrule{2-10}
& \multicolumn{3}{c|}{WMT African} & \multicolumn{3}{c|}{WMT Others} & \multicolumn{3}{c}{WMT Combined}  \\
\midrule
LP & Pearson & Spearman & Kendall & Pearson & Spearman & Kendall & Pearson & Spearman & Kendall \\
\midrule
ary-fra	&	2	&	2	&	2	&	1	&	1	&	1	&	1	&	1	&	1	\\
eng-arz	&	2	&	3	&	3	&	1	&	1	&	1	&	1	&	2	&	2	\\
eng-fra	&	2	&	2	&	2	&	1	&	1	&	1	&	1	&	1	&	1	\\
eng-hau	&	2	&	2	&	2	&	2	&	1	&	1	&	1	&	1	&	1	\\
eng-ibo	&	3	&	2	&	2	&	1	&	1	&	1	&	2	&	1	&	1	\\
eng-kik	&	1	&	2	&	2	&	2	&	1	&	1	&	1	&	1	&	1	\\
eng-luo	&	1	&	1	&	1	&	1	&	1	&	1	&	1	&	1	&	1	\\
eng-som	&	1	&	2	&	2	&	2	&	1	&	1	&	1	&	1	&	1	\\
eng-swh	&	2	&	2	&	2	&	1	&	1	&	1	&	1	&	1	&	1	\\
eng-twi	&	1	&	2	&	2	&	2	&	1	&	2	&	1	&	1	&	1	\\
eng-xho	&	1	&	2	&	2	&	2	&	1	&	1	&	1	&	1	&	1	\\
eng-yor 	&	2	&	2	&	2	&	3	&	1	&	1	&	1	&	1	&	1	\\
eng-yor (it)	&	2	&	1	&	1	&	3	&	1	&	1	&	1	&	1	&	1	\\
eng-yor (movie)	&	2	&	2	&	2	&	3	&	1	&	1	&	1	&	1	&	1	\\
eng-yor (news)	&	2	&	2	&	2	&	1	&	1	&	1	&	1	&	1	&	1	\\
eng-yor (ted)	&	2	&	3	&	3	&	1	&	1	&	1	&	1	&	2	&	2	\\
yor-eng	&	2	&	2	&	2	&	1	&	1	&	1	&	1	&	1	&	1	\\
\midrule
Avg. 	&	1.76	&	2.00	&	2.00	&	1.65	&	\textbf{1.00}	&	\textbf{1.06}	&	\textbf{1.06}	&	1.12	&	1.12	\\
\bottomrule
\end{tabular}
}
\end{center}
\caption{Detailed rankings from the Perm-Input hypothesis test~\citep{deutsch2021statistical} of Pearson, Spearman-rank and Kendall-rank correlation coefficients for MT evaluation models trained with \textbf{single-task learning} based on AfroXLM-Roberta-Large with varied training data settings, corresponding to Table~\ref{tab:afr_xlmr_result}. The averaged ranks are presented in the last row.}
\label{tab:afr_xlmr_stl_rank}
\end{table*}

%% file: tables/afr_xlmr_result_mtl.tex
\begin{table*}[t]
\begin{center}
\small
\centering
\resizebox{2.0\columnwidth}{!}{%
\begin{tabular}{l|ccc|ccc|ccc}
\toprule
& \multicolumn{9}{c}{Training Data Settings} \\
\cmidrule{2-10}
& \multicolumn{3}{c|}{WMT African} & \multicolumn{3}{c|}{WMT Others} & \multicolumn{3}{c}{WMT Combined}  \\
\midrule
LP & Pearson & Spearman & Kendall & Pearson & Spearman & Kendall & Pearson & Spearman & Kendall \\
\midrule
ary-fra	&	0.262	&	0.242	&	0.174	&	0.685	&	0.609	&	0.447	&	0.677	&	0.599	&	0.433	\\
eng-arz	&	0.293	&	0.276	&	0.186	&	0.602	&	0.600	&	0.423	&	0.600	&	0.586	&	0.412	\\
eng-fra	&	0.142	&	0.032	&	0.019	&	0.522	&	0.526	&	0.372	&	0.486	&	0.500	&	0.351	\\
eng-hau	&	0.530	&	0.090	&	0.064	&	0.696	&	0.620	&	0.445	&	0.774	&	0.579	&	0.410	\\
eng-ibo	&	0.124	&	0.196	&	0.140	&	0.649	&	0.616	&	0.445	&	0.621	&	0.507	&	0.364	\\
eng-kik	&	0.519	&	0.233	&	0.161	&	0.523	&	0.410	&	0.276	&	0.630	&	0.332	&	0.225	\\
eng-luo	&	0.320	&	0.270	&	0.181	&	0.433	&	0.359	&	0.251	&	0.460	&	0.370	&	0.252	\\
eng-som	&	0.280	&	0.306	&	0.208	&	0.391	&	0.546	&	0.389	&	0.426	&	0.576	&	0.408	\\
eng-swh	&	0.543	&	0.380	&	0.258	&	0.754	&	0.733	&	0.552	&	0.752	&	0.716	&	0.534	\\
eng-twi	&	0.438	&	0.170	&	0.115	&	0.295	&	0.101	&	0.071	&	0.467	&	0.133	&	0.092	\\
eng-xho	&	0.505	&	0.022	&	0.016	&	0.465	&	0.146	&	0.115	&	0.663	&	0.144	&	0.113	\\
eng-yor (flores)	&	0.716	&	0.186	&	0.126	&	0.694	&	0.365	&	0.256	&	0.811	&	0.323	&	0.227	\\
eng-yor (it)	&	0.741	&	0.298	&	0.208	&	0.659	&	0.418	&	0.298	&	0.817	&	0.261	&	0.255	\\
eng-yor (movie)	&	0.482	&	0.092	&	0.060	&	0.501	&	0.390	&	0.268	&	0.572	&	0.314	&	0.214	\\
eng-yor (news)	&	0.435	&	0.018	&	0.012	&	0.501	&	0.211	&	0.147	&	0.615	&	0.115	&	0.077	\\
eng-yor (ted)	&	0.384	&	0.035	&	0.027	&	0.408	&	0.298	&	0.207	&	0.553	&	0.179	&	0.123	\\
yor-eng	&	0.292	&	0.287	&	0.193	&	0.544	&	0.541	&	0.382	&	0.535	&	0.552	&	0.389	\\
\midrule
Avg.	&	0.412	&	0.184	&	0.126	&	0.548	&	\textbf{0.441}	&	\textbf{0.314}	&	\textbf{0.615}	&	0.399	&	0.287	\\
\bottomrule
\end{tabular}
}
\end{center}
\caption{Correlation coefficients (Pearson, Spearman-rank, Kendall-rank) for MT evaluation models trained with \textbf{multi-task learning} based on AfroXLM-R-L with varied training data settings. Comprehensive results of the Perm-Input hypothesis test~\citep{deutsch2021statistical} are detailed in Table~\ref{tab:afr_xlmr_mtl_rank}. The averaged correlation coefficients are presented in the last row.}
\label{tab:afr_xlmr_result_mlt}
\end{table*}

%% file: tables/afr_xlmr_mtl_rank.tex
\begin{table*}[t]
\begin{center}
\small
\centering
\resizebox{2.0\columnwidth}{!}{%
\begin{tabular}{l|ccc|ccc|ccc}
\toprule
& \multicolumn{9}{c}{Training Data Settings} \\
\cmidrule{2-10}
& \multicolumn{3}{c|}{WMT African} & \multicolumn{3}{c|}{WMT Others} & \multicolumn{3}{c}{WMT Combined}  \\
\midrule
LP & Pearson & Spearman & Kendall & Pearson & Spearman & Kendall & Pearson & Spearman & Kendall \\
\midrule
ary-fra	&	2	&	2	&	2	&	1	&	1	&	1	&	1	&	1	&	1	\\
eng-arz	&	2	&	2	&	2	&	1	&	1	&	1	&	1	&	1	&	1	\\
eng-fra	&	3	&	3	&	2	&	1	&	1	&	1	&	2	&	2	&	1	\\
eng-hau	&	2	&	2	&	3	&	1	&	1	&	1	&	1	&	1	&	2	\\
eng-ibo	&	2	&	3	&	3	&	1	&	1	&	1	&	1	&	2	&	2	\\
eng-kik	&	2	&	3	&	2	&	2	&	1	&	1	&	1	&	2	&	2	\\
eng-luo	&	2	&	2	&	2	&	1	&	1	&	1	&	1	&	1	&	1	\\
eng-som	&	3	&	3	&	2	&	2	&	2	&	1	&	1	&	1	&	1	\\
eng-swh	&	2	&	2	&	2	&	1	&	1	&	1	&	1	&	1	&	1	\\
eng-twi	&	1	&	1	&	1	&	2	&	1	&	1	&	1	&	1	&	1	\\
eng-xho	&	2	&	2	&	2	&	2	&	1	&	1	&	1	&	1	&	1	\\
eng-yor 	&	2	&	2	&	2	&	2	&	1	&	1	&	1	&	1	&	1	\\
eng-yor (it)	&	2	&	2	&	2	&	3	&	1	&	1	&	1	&	1	&	2	\\
eng-yor (movie)	&	2	&	3	&	3	&	2	&	1	&	1	&	1	&	2	&	2	\\
eng-yor (news)	&	2	&	2	&	2	&	2	&	1	&	1	&	1	&	2	&	2	\\
eng-yor (ted)	&	2	&	3	&	3	&	1	&	1	&	1	&	1	&	2	&	2	\\
yor-eng	&	2	&	2	&	2	&	1	&	1	&	1	&	1	&	1	&	1	\\
\midrule
Avg. 	&	2.06	&	2.29	&	2.18	&	1.53	&	\textbf{1.06}	&	\textbf{1.00}	&	\textbf{1.06}	&	1.35	&	1.41	\\
\bottomrule
\end{tabular}
}
\end{center}
\caption{Detailed rankings from the Perm-Input hypothesis test~\citep{deutsch2021statistical} of Pearson, Spearman-rank and Kendall-rank correlation coefficients for MT evaluation models trained with \textbf{multi-task learning} based on AfroXLM-Roberta-Large with varied training data settings, corresponding to Table~\ref{tab:afr_xlmr_result_mlt}. The averaged ranks are presented in the last row.}
\label{tab:afr_xlmr_mtl_rank}
\end{table*}

%% file: tables/ref_free_rank_result.tex
\begin{table*}[!t]
\begin{center}
\small
\centering
\resizebox{2.0\columnwidth}{!}{%
\begin{tabular}{l|cc|cc|cc|cc|cc}
\toprule
& \multicolumn{2}{c|}{\textbf{LLM Prompting}} 
& \multicolumn{8}{c}{\textbf{Learned refence-free QE Metric}} \\
\cmidrule{2-11}
& \multicolumn{2}{c|}{\multirow{2}{*}{GPT4}} 
& \multicolumn{2}{c|}{\textbf{Baseline}}
& \multicolumn{4}{c|}{\textbf{Single Task} (Ours)}
& \multicolumn{2}{c}{\textbf{Multi Task} (Ours)} \\
\cmidrule{4-11}
& & & \multicolumn{2}{c|}{CometKiwi}
& \multicolumn{2}{c|}{InfoXLM-L}
& \multicolumn{2}{c|}{AfroXLM-R-L \ding{72}}
& \multicolumn{2}{c}{AfroXLM-R-L \ding{72}}  \\
\cmidrule{2-11}
LP & Pearson & Spearman & Pearson & Spearman & Pearson & Spearman & Pearson & Spearman & Pearson & Spearman \\
\midrule
ary-fra	&	1	&	1	&	2	&	2	&	2	&	1	&	3	&	2	&	1	&	1	\\
eng-arz	&	2	&	1	&	1	&	1	&	2	&	2	&	2	&	2	&	1	&	1	\\
eng-fra	&	1	&	1	&	1	&	1	&	2	&	1	&	1	&	1	&	1	&	1	\\
eng-hau	&	2	&	3	&	2	&	3	&	2	&	3	&	1	&	2	&	1	&	1	\\
eng-ibo	&	3	&	3	&	3	&	3	&	2	&	2	&	1	&	1	&	1	&	1	\\
eng-kik	&	4	&	2	&	3	&	2	&	2	&	1	&	1	&	1	&	2	&	2	\\
eng-luo	&	3	&	2	&	2	&	1	&	3	&	1	&	1	&	1	&	2	&	1	\\
eng-som	&	2	&	3	&	1	&	2	&	2	&	3	&	1	&	1	&	1	&	1	\\
eng-swh	&	2	&	1	&	1	&	1	&	2	&	2	&	3	&	2	&	2	&	1	\\
eng-twi	&	1	&	1	&	2	&	1	&	2	&	2	&	1	&	1	&	1	&	1	\\
eng-xho	&	1	&	1	&	2	&	2	&	2	&	1	&	1	&	1	&	1	&	1	\\
eng-yor 	&	4	&	1	&	4	&	2	&	3	&	2	&	1	&	1	&	2	&	2	\\
eng-yor (it)	&	2	&	2	&	2	&	1	&	2	&	2	&	1	&	2	&	1	&	1	\\
eng-yor (movie)	&	2	&	1	&	4	&	3	&	3	&	2	&	1	&	2	&	2	&	2	\\
eng-yor (news)	&	2	&	1	&	2	&	2	&	2	&	2	&	1	&	1	&	1	&	1	\\
eng-yor (ted)	&	1	&	1	&	2	&	1	&	2	&	1	&	1	&	1	&	1	&	1	\\
yor-eng	&	1	&	1	&	3	&	2	&	3	&	3	&	2	&	2	&	1	&	1	\\
\midrule
Avg.	&	2.00	&	1.53	&	2.18	&	1.76	&	2.24	&	1.82	&	\textbf{1.35}	&	\textbf{1.41}	&	\textbf{1.29}	&	\textbf{1.18}	\\
\bottomrule
\end{tabular}
}
\end{center}
\caption{Detailed rankings from the Perm-Input hypothesis test~\citep{deutsch2021statistical} of Pearson and Spearman-rank correlation coefficients corresponding to Table~\ref{tab:ref_free_result}. The averaged ranks are presented in the last row.}
\label{tab:ref_free_rank}
\end{table*}

%% file: tables/wmt_afr_eval.tex
\begin{table*}[!t]
\small
\begin{center}
\centering
\resizebox{1.0\columnwidth}{!}{%
\begin{tabular}{l|ccc}
\toprule
\multicolumn{4}{c}{\textbf{MT Evaluation}} \\
\midrule
\textbf{MT Evaluation System} & Pearson & Spearman & Kendall \\
\midrule
COMET22~\citep{rei2022comet}	& 0.578	& 0.482	& 0.332 \\
AfriCOMET-STL (Ours)	& \textbf{0.618}	& \textbf{0.507}	& \textbf{0.351} \\
AfriCOMET-MTL (Ours)	& 0.591	& 0.486	& 0.333 \\
\midrule
\midrule
\multicolumn{4}{c}{\textbf{Quality Estimation}} \\
\midrule
\textbf{QE System} & Pearson & Spearman & Kendall \\
\midrule
CometKiwi~\citep{rei-etal-2022-cometkiwi}	& 0.242	& 0.219	& - \\
AfriCOMET-QE-STL (Ours)	& 0.552	& 0.413	& - \\
AfriCOMET-QE-MTL (Ours)	& \textbf{0.558}	& \textbf{0.445}	& - \\
\bottomrule
\end{tabular}
}
\end{center}
\caption{Performance of COMET22, AfriCOMET, CometKiwi and AfriCOMET-QE on the ``\textbf{WMT African}'' dataset, a human evaluation set from the WMT 2022 shared task: ``Large-Scale Machine Translation Evaluation for African Languages''~\citep{adelani-etal-2022-findings}. Results are reported in terms of correlation coefficients: Pearson, Spearman-rank, and Kendall-rank for MT evaluation; Pearson and Spearman-rank for QE. MT evaluation systems are evaluated using the source, the machine translation, and the reference as model inputs, while QE systems are assessed relying only on the source and the machine translation. Correlations are calculated between human-annotated DA scores and automatic scores.}
\label{tab:wmt_afr_eval_result}
\end{table*}

%% file: tables/wmt22-qe-eng-yor.tex
\begin{table*}[!t]
\small
\begin{center}
\centering
\resizebox{1.0\columnwidth}{!}{%
\begin{tabular}{l|cc}
\toprule
\multicolumn{3}{c}{\textbf{Quality Estimation}} \\
\midrule
\textbf{QE System} & Pearson & Spearman	\\
\midrule
CometKiwi~\citep{rei-etal-2022-cometkiwi} &	0.153 &	0.118 \\
AfriCOMET-QE-STL (Ours) &	0.461 & 0.482 \\
AfriCOMET-QE-MTL (Ours) &	\textbf{0.485} &	\textbf{0.495} \\
\bottomrule
\end{tabular}
}
\end{center}
\caption{Performance of CometKiwi and our benchmark AfriCOMET-QE systems on the English-Yoruba test set (\url{https://github.com/WMT-QE-Task/wmt-qe-2022-data/tree/main/test_data-gold_labels/task1_da/en-yo}) from the WMT 2022 Quality Estimation Shared Task~\citep{zerva2022findings}. This dataset includes 1010 DA annotations. Results are reported in terms of Pearson and Spearman-rank correlation coefficients. All metrics are trained on the ``WMT Others'' dataset, and they are evaluated with source and machine translation as model inputs. Correlations are calculated between human-annotated DA scores and automatic scores.}

\label{tab:wmt22-qe-eng-yor-result}
\end{table*}

%% file: tables/generalization.tex
\begin{table*}[!t]
\centering
\scalebox{0.5}{
\begin{tabular}{l|ccc|ccc|ccc|cc|cc|cc}
\toprule
& \multicolumn{9}{c|}{\textbf{MT Evaluation}} 
& \multicolumn{6}{c}{\textbf{Quality Estimation}} \\
\cmidrule{2-16}
& \multicolumn{3}{c|}{\textbf{COMET22}}
& \multicolumn{3}{c|}{\textbf{AfriCOMET-STL} (Ours)}
& \multicolumn{3}{c|}{\textbf{AfriCOMET-MTL} (Ours)}
& \multicolumn{2}{c|}{\textbf{CometKiwi}}
& \multicolumn{2}{c|}{\textbf{AfriCOMET-QE-STL} (Ours)}
& \multicolumn{2}{c}{\textbf{AfriCOMET-QE-MTL} (Ours)} \\
\cmidrule{2-16}
LP 
& Pearson & Spearman & Kendall 
& Pearson & Spearman & Kendall 
& Pearson & Spearman & Kendall 
& Pearson & Spearman
& Pearson & Spearman
& Pearson & Spearman \\
\midrule
eng-deu	&	0.312	&	0.319	&	0.244	&	0.263	&	0.277	&	0.211	&	0.265	&	0.286	&	0.219	&	0.254	&	0.273	&	0.264	&	0.256	&	0.228	&	0.247	\\
eng-rus	&	0.361	&	0.370	&	0.286	&	0.344	&	0.341	&	0.264	&	0.381	&	0.380	&	0.295	&	0.357	&	0.360	&	0.326	&	0.337	&	0.336	&	0.358	\\
zho-eng	&	0.428	&	0.490	&	0.357	&	0.427	&	0.487	&	0.355	&	0.420	&	0.479	&	0.348	&	0.362	&	0.423	&	0.370	&	0.421	&	0.367	&	0.431	\\
\midrule
Avg.	&	\textbf{0.367}	&	\textbf{0.393}	&	\textbf{0.296}	&	0.345	&	0.368	&	0.277	&	0.355	&	0.382	&	0.287	&	\textbf{0.324}	&	\textbf{0.352}	&	0.320	&	0.338	&	0.310	&	0.345	\\

\bottomrule
\end{tabular}
}
\caption{Generalization assessments: performance of COMET22 and AfriCOMET for MT evaluation tasks, and performance of CometKiwi and AfriCOMET-QE for QE tasks, on the English-German (eng-deu), English-Russian (eng-rus), and Chinese-English (zho-eng) MQM datasets from the WMT 2022 News Domain Translation Shared Task (\url{https://github.com/google/wmt-mqm-human-evaluation}). These three datasets serve as validation sets for COMET22 and CometKiwi, while remaining unseen in either training or validation for AfriCOMET and AfriCOMET-QE.}
\label{tab:generalization_result}
\end{table*}

%% file: main.bbl
\begin{thebibliography}{47}
\expandafter\ifx\csname natexlab\endcsname\relax\def\natexlab#1{#1}\fi

\bibitem[{Adelani et~al.(2022)Adelani, Alam, Anastasopoulos, Bhagia, Costa-juss{\`a}, Dodge, Faisal, Federmann, Fedorova, Guzm{\'a}n, Koshelev, Maillard, Marivate, Mbuya, Mourachko, Saleem, Schwenk, and Wenzek}]{adelani-etal-2022-findings}
David Adelani, Md~Mahfuz~Ibn Alam, Antonios Anastasopoulos, Akshita Bhagia, Marta~R. Costa-juss{\`a}, Jesse Dodge, Fahim Faisal, Christian Federmann, Natalia Fedorova, Francisco Guzm{\'a}n, Sergey Koshelev, Jean Maillard, Vukosi Marivate, Jonathan Mbuya, Alexandre Mourachko, Safiyyah Saleem, Holger Schwenk, and Guillaume Wenzek. 2022.
\newblock \href {https://aclanthology.org/2022.wmt-1.72} {Findings of the {WMT}{'}22 shared task on large-scale machine translation evaluation for {A}frican languages}.
\newblock In \emph{Proceedings of the Seventh Conference on Machine Translation (WMT)}, pages 773--800, Abu Dhabi, United Arab Emirates (Hybrid). Association for Computational Linguistics.

\bibitem[{Adelani et~al.(2021)Adelani, Ruiter, Alabi, Adebonojo, Ayeni, Adeyemi, Awokoya, and Espa{\~n}a-Bonet}]{adelani2021effect}
David~I Adelani, Dana Ruiter, Jesujoba~O Alabi, Damilola Adebonojo, Adesina Ayeni, Mofe Adeyemi, Ayodele Awokoya, and Cristina Espa{\~n}a-Bonet. 2021.
\newblock The effect of domain and diacritics in yor$\backslash$ub$\backslash$'a-english neural machine translation.
\newblock \emph{arXiv preprint arXiv:2103.08647}.

\bibitem[{Alabi et~al.(2022)Alabi, Adelani, Mosbach, and Klakow}]{alabi-etal-2022-adapting}
Jesujoba~O. Alabi, David~Ifeoluwa Adelani, Marius Mosbach, and Dietrich Klakow. 2022.
\newblock \href {https://aclanthology.org/2022.coling-1.382} {Adapting pre-trained language models to {A}frican languages via multilingual adaptive fine-tuning}.
\newblock In \emph{Proceedings of the 29th International Conference on Computational Linguistics}, pages 4336--4349, Gyeongju, Republic of Korea. International Committee on Computational Linguistics.

\bibitem[{Banerjee and Lavie(2005)}]{banerjee-lavie-2005-meteor}
Satanjeev Banerjee and Alon Lavie. 2005.
\newblock \href {https://aclanthology.org/W05-0909} {{METEOR}: An automatic metric for {MT} evaluation with improved correlation with human judgments}.
\newblock In \emph{Proceedings of the {ACL} Workshop on Intrinsic and Extrinsic Evaluation Measures for Machine Translation and/or Summarization}, pages 65--72, Ann Arbor, Michigan. Association for Computational Linguistics.

\bibitem[{Bapna et~al.(2022)Bapna, Caswell, Kreutzer, Firat, van Esch, Siddhant, Niu, Baljekar, Garc{\'i}a, Macherey, Breiner, Axelrod, Riesa, Cao, Chen, Macherey, Krikun, Wang, Gutkin, Shah, Huang, Chen, Wu, and Hughes}]{Bapna2022BuildingMT}
Ankur Bapna, Isaac Caswell, Julia Kreutzer, Orhan Firat, Daan van Esch, Aditya Siddhant, Mengmeng Niu, Pallavi~N. Baljekar, Xavier Garc{\'i}a, Wolfgang Macherey, Theresa Breiner, Vera Axelrod, Jason Riesa, Yuan Cao, Mia~Xu Chen, Klaus Macherey, Maxim Krikun, Pidong Wang, Alexander Gutkin, Apurva Shah, Yanping Huang, Z.~Chen, Yonghui Wu, and Macduff Hughes. 2022.
\newblock \href {https://api.semanticscholar.org/CorpusID:248572135} {Building machine translation systems for the next thousand languages}.
\newblock \emph{ArXiv}, abs/2205.03983.

\bibitem[{Bentivogli et~al.(2018)Bentivogli, Cettolo, Federico, and Christian}]{bentivogli2018machine}
Luisa Bentivogli, Mauro Cettolo, Marcello Federico, and Federmann Christian. 2018.
\newblock Machine translation human evaluation: an investigation of evaluation based on post-editing and its relation with direct assessment.
\newblock In \emph{Proceedings of the 15th International Workshop on Spoken Language Translation (IWSLT 2018)}, pages 62--69.

\bibitem[{Chatzikoumi(2020)}]{chatzikoumi2020evaluate}
Eirini Chatzikoumi. 2020.
\newblock How to evaluate machine translation: A review of automated and human metrics.
\newblock \emph{Natural Language Engineering}, 26(2):137--161.

\bibitem[{Chi et~al.(2020)Chi, Dong, Wei, Yang, Singhal, Wang, Song, Mao, Huang, and Zhou}]{chi2020infoxlm}
Zewen Chi, Li~Dong, Furu Wei, Nan Yang, Saksham Singhal, Wenhui Wang, Xia Song, Xian-Ling Mao, Heyan Huang, and Ming Zhou. 2020.
\newblock Infoxlm: An information-theoretic framework for cross-lingual language model pre-training.
\newblock \emph{arXiv preprint arXiv:2007.07834}.

\bibitem[{Conneau et~al.(2019)Conneau, Khandelwal, Goyal, Chaudhary, Wenzek, Guzm{\'a}n, Grave, Ott, Zettlemoyer, and Stoyanov}]{conneau2019unsupervised}
Alexis Conneau, Kartikay Khandelwal, Naman Goyal, Vishrav Chaudhary, Guillaume Wenzek, Francisco Guzm{\'a}n, Edouard Grave, Myle Ott, Luke Zettlemoyer, and Veselin Stoyanov. 2019.
\newblock Unsupervised cross-lingual representation learning at scale.
\newblock \emph{arXiv preprint arXiv:1911.02116}.

\bibitem[{Conneau et~al.(2020)Conneau, Khandelwal, Goyal, Chaudhary, Wenzek, Guzm{\'a}n, Grave, Ott, Zettlemoyer, and Stoyanov}]{conneau-etal-2020-unsupervised}
Alexis Conneau, Kartikay Khandelwal, Naman Goyal, Vishrav Chaudhary, Guillaume Wenzek, Francisco Guzm{\'a}n, Edouard Grave, Myle Ott, Luke Zettlemoyer, and Veselin Stoyanov. 2020.
\newblock \href {https://doi.org/10.18653/v1/2020.acl-main.747} {Unsupervised cross-lingual representation learning at scale}.
\newblock In \emph{Proceedings of the 58th Annual Meeting of the Association for Computational Linguistics}, pages 8440--8451, Online. Association for Computational Linguistics.

\bibitem[{Deutsch et~al.(2021)Deutsch, Dror, and Roth}]{deutsch2021statistical}
Daniel Deutsch, Rotem Dror, and Dan Roth. 2021.
\newblock A statistical analysis of summarization evaluation metrics using resampling methods.
\newblock \emph{Transactions of the Association for Computational Linguistics}, 9:1132--1146.

\bibitem[{Deutsch et~al.(2023)Deutsch, Foster, and Freitag}]{deutsch2023ties}
Daniel Deutsch, George Foster, and Markus Freitag. 2023.
\newblock Ties matter: Meta-evaluating modern metrics with pairwise accuracy and tie calibration.
\newblock In \emph{Proceedings of the 2023 Conference on Empirical Methods in Natural Language Processing}, pages 12914--12929.

\bibitem[{Fan et~al.(2021{\natexlab{a}})Fan, Bhosale, Schwenk, Ma, El-Kishky, Goyal, Baines, Celebi, Wenzek, Chaudhary, Goyal, Birch, Liptchinsky, Edunov, Grave, Auli, and Joulin}]{m2m100_fan}
Angela Fan, Shruti Bhosale, Holger Schwenk, Zhiyi Ma, Ahmed El-Kishky, Siddharth Goyal, Mandeep Baines, Onur Celebi, Guillaume Wenzek, Vishrav Chaudhary, Naman Goyal, Tom Birch, Vitaliy Liptchinsky, Sergey Edunov, Edouard Grave, Michael Auli, and Armand Joulin. 2021{\natexlab{a}}.
\newblock Beyond english-centric multilingual machine translation.
\newblock \emph{J. Mach. Learn. Res.}, 22(1).

\bibitem[{Fan et~al.(2021{\natexlab{b}})Fan, Bhosale, Schwenk, Ma, El-Kishky, Goyal, Baines, Celebi, Wenzek, Chaudhary et~al.}]{fan2021beyond}
Angela Fan, Shruti Bhosale, Holger Schwenk, Zhiyi Ma, Ahmed El-Kishky, Siddharth Goyal, Mandeep Baines, Onur Celebi, Guillaume Wenzek, Vishrav Chaudhary, et~al. 2021{\natexlab{b}}.
\newblock Beyond english-centric multilingual machine translation.
\newblock \emph{The Journal of Machine Learning Research}, 22(1):4839--4886.

\bibitem[{Fan et~al.(2019)Fan, Wang, Li, Zhou, Chen, and Si}]{fan2019bilingual}
Kai Fan, Jiayi Wang, Bo~Li, Fengming Zhou, Boxing Chen, and Luo Si. 2019.
\newblock “bilingual expert” can find translation errors.
\newblock In \emph{Proceedings of the AAAI Conference on Artificial Intelligence}, volume~33, pages 6367--6374.

\bibitem[{Fomicheva et~al.(2021)Fomicheva, Lertvittayakumjorn, Zhao, Eger, and Gao}]{fomicheva-etal-2021-eval4nlp}
Marina Fomicheva, Piyawat Lertvittayakumjorn, Wei Zhao, Steffen Eger, and Yang Gao. 2021.
\newblock \href {https://doi.org/10.18653/v1/2021.eval4nlp-1.17} {The {E}val4{NLP} shared task on explainable quality estimation: Overview and results}.
\newblock In \emph{Proceedings of the 2nd Workshop on Evaluation and Comparison of NLP Systems}, pages 165--178, Punta Cana, Dominican Republic. Association for Computational Linguistics.

\bibitem[{Fomicheva et~al.(2020)Fomicheva, Sun, Fonseca, Zerva, Blain, Chaudhary, Guzm{\'a}n, Lopatina, Specia, and Martins}]{fomicheva2020mlqe}
Marina Fomicheva, Shuo Sun, Erick Fonseca, Chrysoula Zerva, Fr{\'e}d{\'e}ric Blain, Vishrav Chaudhary, Francisco Guzm{\'a}n, Nina Lopatina, Lucia Specia, and Andr{\'e}~FT Martins. 2020.
\newblock Mlqe-pe: A multilingual quality estimation and post-editing dataset.
\newblock \emph{arXiv preprint arXiv:2010.04480}.

\bibitem[{Freitag et~al.(2021{\natexlab{a}})Freitag, Foster, Grangier, Ratnakar, Tan, and Macherey}]{freitag2021experts}
Markus Freitag, George Foster, David Grangier, Viresh Ratnakar, Qijun Tan, and Wolfgang Macherey. 2021{\natexlab{a}}.
\newblock Experts, errors, and context: A large-scale study of human evaluation for machine translation.
\newblock \emph{Transactions of the Association for Computational Linguistics}, 9:1460--1474.

\bibitem[{Freitag et~al.(2022)Freitag, Rei, Mathur, Lo, Stewart, Avramidis, Kocmi, Foster, Lavie, and Martins}]{freitag-etal-2022-results}
Markus Freitag, Ricardo Rei, Nitika Mathur, Chi-kiu Lo, Craig Stewart, Eleftherios Avramidis, Tom Kocmi, George Foster, Alon Lavie, and Andr{\'e} F.~T. Martins. 2022.
\newblock \href {https://aclanthology.org/2022.wmt-1.2} {Results of {WMT}22 metrics shared task: Stop using {BLEU} {--} neural metrics are better and more robust}.
\newblock In \emph{Proceedings of the Seventh Conference on Machine Translation (WMT)}, pages 46--68, Abu Dhabi, United Arab Emirates (Hybrid). Association for Computational Linguistics.

\bibitem[{Freitag et~al.(2021{\natexlab{b}})Freitag, Rei, Mathur, Lo, Stewart, Foster, Lavie, and Bojar}]{freitag2021results}
Markus Freitag, Ricardo Rei, Nitika Mathur, Chi-kiu Lo, Craig Stewart, George Foster, Alon Lavie, and Ond{\v{r}}ej Bojar. 2021{\natexlab{b}}.
\newblock Results of the wmt21 metrics shared task: Evaluating metrics with expert-based human evaluations on ted and news domain.
\newblock In \emph{Proceedings of the Sixth Conference on Machine Translation}, pages 733--774.

\bibitem[{Goyal et~al.(2022)Goyal, Gao, Chaudhary, Chen, Wenzek, Ju, Krishnan, Ranzato, Guzm{\'a}n, and Fan}]{goyal2022flores}
Naman Goyal, Cynthia Gao, Vishrav Chaudhary, Peng-Jen Chen, Guillaume Wenzek, Da~Ju, Sanjana Krishnan, Marc’Aurelio Ranzato, Francisco Guzm{\'a}n, and Angela Fan. 2022.
\newblock The flores-101 evaluation benchmark for low-resource and multilingual machine translation.
\newblock \emph{Transactions of the Association for Computational Linguistics}, 10:522--538.

\bibitem[{Graham et~al.(2013)Graham, Baldwin, Moffat, and Zobel}]{graham-etal-2013-continuous}
Yvette Graham, Timothy Baldwin, Alistair Moffat, and Justin Zobel. 2013.
\newblock \href {https://aclanthology.org/W13-2305} {Continuous measurement scales in human evaluation of machine translation}.
\newblock In \emph{Proceedings of the 7th Linguistic Annotation Workshop and Interoperability with Discourse}, pages 33--41, Sofia, Bulgaria. Association for Computational Linguistics.

\bibitem[{Graham et~al.(2017)Graham, Baldwin, Moffat, and Zobel}]{graham2017can}
Yvette Graham, Timothy Baldwin, Alistair Moffat, and Justin Zobel. 2017.
\newblock Can machine translation systems be evaluated by the crowd alone.
\newblock \emph{Natural Language Engineering}, 23(1):3--30.

\bibitem[{Kocmi et~al.(2023)Kocmi, Avramidis, Bawden, Bojar, Dvorkovich, Federmann, Fishel, Freitag, Gowda, Grundkiewicz et~al.}]{kocmi2023findings}
Tom Kocmi, Eleftherios Avramidis, Rachel Bawden, Ond{\v{r}}ej Bojar, Anton Dvorkovich, Christian Federmann, Mark Fishel, Markus Freitag, Thamme Gowda, Roman Grundkiewicz, et~al. 2023.
\newblock Findings of the 2023 conference on machine translation (wmt23): Llms are here but not quite there yet.
\newblock In \emph{Proceedings of the Eighth Conference on Machine Translation}, pages 1--42.

\bibitem[{Kudugunta et~al.(2023)Kudugunta, Caswell, Zhang, Garc{\'i}a, Choquette-Choo, Lee, Xin, Kusupati, Stella, Bapna, and Firat}]{Kudugunta2023MADLAD400AM}
Sneha Kudugunta, Isaac Caswell, Biao Zhang, Xavier Garc{\'i}a, Christopher~A. Choquette-Choo, Katherine Lee, Derrick Xin, Aditya Kusupati, Romi Stella, Ankur Bapna, and Orhan Firat. 2023.
\newblock \href {https://api.semanticscholar.org/CorpusID:261682406} {Madlad-400: A multilingual and document-level large audited dataset}.
\newblock \emph{ArXiv}, abs/2309.04662.

\bibitem[{Lommel et~al.(2014)Lommel, Uszkoreit, and Burchardt}]{Lommel2014MultidimensionalQM}
Arle Lommel, Hans Uszkoreit, and Aljoscha Burchardt. 2014.
\newblock \href {https://api.semanticscholar.org/CorpusID:55606096} {Multidimensional quality metrics (mqm): A framework for declaring and describing translation quality metrics}.
\newblock In \emph{Tradumàtica}.

\bibitem[{Nekoto et~al.(2020)Nekoto, Marivate, Matsila, Fasubaa, Fagbohungbe, Akinola, Muhammad, Kabongo~Kabenamualu, Osei, Sackey, Niyongabo, Macharm, Ogayo, Ahia, Berhe, Adeyemi, Mokgesi-Selinga, Okegbemi, Martinus, Tajudeen, Degila, Ogueji, Siminyu, Kreutzer, Webster, Ali, Abbott, Orife, Ezeani, Dangana, Kamper, Elsahar, Duru, Kioko, Espoir, van Biljon, Whitenack, Onyefuluchi, Emezue, Dossou, Sibanda, Bassey, Olabiyi, Ramkilowan, {\"O}ktem, Akinfaderin, and Bashir}]{nekoto-etal-2020-participatory}
Wilhelmina Nekoto, Vukosi Marivate, Tshinondiwa Matsila, Timi Fasubaa, Taiwo Fagbohungbe, Solomon~Oluwole Akinola, Shamsuddeen Muhammad, Salomon Kabongo~Kabenamualu, Salomey Osei, Freshia Sackey, Rubungo~Andre Niyongabo, Ricky Macharm, Perez Ogayo, Orevaoghene Ahia, Musie~Meressa Berhe, Mofetoluwa Adeyemi, Masabata Mokgesi-Selinga, Lawrence Okegbemi, Laura Martinus, Kolawole Tajudeen, Kevin Degila, Kelechi Ogueji, Kathleen Siminyu, Julia Kreutzer, Jason Webster, Jamiil~Toure Ali, Jade Abbott, Iroro Orife, Ignatius Ezeani, Idris~Abdulkadir Dangana, Herman Kamper, Hady Elsahar, Goodness Duru, Ghollah Kioko, Murhabazi Espoir, Elan van Biljon, Daniel Whitenack, Christopher Onyefuluchi, Chris~Chinenye Emezue, Bonaventure F.~P. Dossou, Blessing Sibanda, Blessing Bassey, Ayodele Olabiyi, Arshath Ramkilowan, Alp {\"O}ktem, Adewale Akinfaderin, and Abdallah Bashir. 2020.
\newblock \href {https://doi.org/10.18653/v1/2020.findings-emnlp.195} {Participatory research for low-resourced machine translation: A case study in {A}frican languages}.
\newblock In \emph{Findings of the Association for Computational Linguistics: EMNLP 2020}, pages 2144--2160, Online. Association for Computational Linguistics.

\bibitem[{{NLLB-Team} et~al.(2022){NLLB-Team}, Costa-juss{\`a}, Cross, cCelebi, Elbayad, Heafield, Heffernan, Kalbassi, Lam, Licht, Maillard, Sun, Wang, Wenzek, Youngblood, Akula, Barrault, Gonzalez, Hansanti, Hoffman, Jarrett, Sadagopan, Rowe, Spruit, Tran, Andrews, Ayan, Bhosale, Edunov, Fan, Gao, Goswami, Guzm'an, Koehn, Mourachko, Ropers, Saleem, Schwenk, and Wang}]{team2022NoLL}
{NLLB-Team}, Marta~Ruiz Costa-juss{\`a}, James Cross, Onur cCelebi, Maha Elbayad, Kenneth Heafield, Kevin Heffernan, Elahe Kalbassi, Janice Lam, Daniel Licht, Jean Maillard, Anna Sun, Skyler Wang, Guillaume Wenzek, Alison Youngblood, Bapi Akula, Lo{\"i}c Barrault, Gabriel~Mejia Gonzalez, Prangthip Hansanti, John Hoffman, Semarley Jarrett, Kaushik~Ram Sadagopan, Dirk Rowe, Shannon~L. Spruit, C.~Tran, Pierre~Yves Andrews, Necip~Fazil Ayan, Shruti Bhosale, Sergey Edunov, Angela Fan, Cynthia Gao, Vedanuj Goswami, Francisco Guzm'an, Philipp Koehn, Alexandre Mourachko, Christophe Ropers, Safiyyah Saleem, Holger Schwenk, and Jeff Wang. 2022.
\newblock \href {https://api.semanticscholar.org/CorpusID:250425961} {No language left behind: Scaling human-centered machine translation}.
\newblock \emph{ArXiv}, abs/2207.04672.

\bibitem[{Papineni et~al.(2002)Papineni, Roukos, Ward, and Zhu}]{papineni-etal-2002-bleu}
Kishore Papineni, Salim Roukos, Todd Ward, and Wei-Jing Zhu. 2002.
\newblock \href {https://doi.org/10.3115/1073083.1073135} {{B}leu: a method for automatic evaluation of machine translation}.
\newblock In \emph{Proceedings of the 40th Annual Meeting of the Association for Computational Linguistics}, pages 311--318, Philadelphia, Pennsylvania, USA. Association for Computational Linguistics.

\bibitem[{Pavlick and Tetreault(2016)}]{pavlick-tetreault-2016-empirical}
Ellie Pavlick and Joel Tetreault. 2016.
\newblock \href {https://doi.org/10.1162/tacl_a_00083} {An empirical analysis of formality in online communication}.
\newblock \emph{Transactions of the Association for Computational Linguistics}, 4:61--74.

\bibitem[{Popovi{\'c}(2015)}]{popovic-2015-chrf}
Maja Popovi{\'c}. 2015.
\newblock \href {https://doi.org/10.18653/v1/W15-3049} {chr{F}: character n-gram {F}-score for automatic {MT} evaluation}.
\newblock In \emph{Proceedings of the Tenth Workshop on Statistical Machine Translation}, pages 392--395, Lisbon, Portugal. Association for Computational Linguistics.

\bibitem[{Popovi{\'c}(2017)}]{popovic-2017-chrf}
Maja Popovi{\'c}. 2017.
\newblock \href {https://doi.org/10.18653/v1/W17-4770} {chr{F}++: words helping character n-grams}.
\newblock In \emph{Proceedings of the Second Conference on Machine Translation}, pages 612--618, Copenhagen, Denmark. Association for Computational Linguistics.

\bibitem[{Post(2018)}]{post-2018-call}
Matt Post. 2018.
\newblock \href {https://doi.org/10.18653/v1/W18-6319} {A call for clarity in reporting {BLEU} scores}.
\newblock In \emph{Proceedings of the Third Conference on Machine Translation: Research Papers}, pages 186--191, Brussels, Belgium. Association for Computational Linguistics.

\bibitem[{Ranasinghe et~al.(2020)Ranasinghe, Orasan, and Mitkov}]{ranasinghe2020transquest}
Tharindu Ranasinghe, Constantin Orasan, and Ruslan Mitkov. 2020.
\newblock Transquest: Translation quality estimation with cross-lingual transformers.
\newblock \emph{arXiv preprint arXiv:2011.01536}.

\bibitem[{Rei et~al.(2022{\natexlab{a}})Rei, De~Souza, Alves, Zerva, Farinha, Glushkova, Lavie, Coheur, and Martins}]{rei2022comet}
Ricardo Rei, Jos{\'e}~GC De~Souza, Duarte Alves, Chrysoula Zerva, Ana~C Farinha, Taisiya Glushkova, Alon Lavie, Luisa Coheur, and Andr{\'e}~FT Martins. 2022{\natexlab{a}}.
\newblock Comet-22: Unbabel-ist 2022 submission for the metrics shared task.
\newblock In \emph{Proceedings of the Seventh Conference on Machine Translation (WMT)}, pages 578--585.

\bibitem[{Rei et~al.(2020)Rei, Stewart, Farinha, and Lavie}]{rei-etal-2020-comet}
Ricardo Rei, Craig Stewart, Ana~C Farinha, and Alon Lavie. 2020.
\newblock \href {https://doi.org/10.18653/v1/2020.emnlp-main.213} {{COMET}: A neural framework for {MT} evaluation}.
\newblock In \emph{Proceedings of the 2020 Conference on Empirical Methods in Natural Language Processing (EMNLP)}, pages 2685--2702, Online. Association for Computational Linguistics.

\bibitem[{Rei et~al.(2022{\natexlab{b}})Rei, Treviso, Guerreiro, Zerva, Farinha, Maroti, C.~de Souza, Glushkova, Alves, Coheur, Lavie, and Martins}]{rei-etal-2022-cometkiwi}
Ricardo Rei, Marcos Treviso, Nuno~M. Guerreiro, Chrysoula Zerva, Ana~C Farinha, Christine Maroti, Jos{\'e}~G. C.~de Souza, Taisiya Glushkova, Duarte Alves, Luisa Coheur, Alon Lavie, and Andr{\'e} F.~T. Martins. 2022{\natexlab{b}}.
\newblock \href {https://aclanthology.org/2022.wmt-1.60} {{C}omet{K}iwi: {IST}-unbabel 2022 submission for the quality estimation shared task}.
\newblock In \emph{Proceedings of the Seventh Conference on Machine Translation (WMT)}, pages 634--645, Abu Dhabi, United Arab Emirates (Hybrid). Association for Computational Linguistics.

\bibitem[{Sai~B et~al.(2023)Sai~B, Dixit, Nagarajan, Kunchukuttan, Kumar, Khapra, and Dabre}]{sai-b-etal-2023-indicmt}
Ananya Sai~B, Tanay Dixit, Vignesh Nagarajan, Anoop Kunchukuttan, Pratyush Kumar, Mitesh~M. Khapra, and Raj Dabre. 2023.
\newblock \href {https://doi.org/10.18653/v1/2023.acl-long.795} {{I}ndic{MT} eval: A dataset to meta-evaluate machine translation metrics for {I}ndian languages}.
\newblock In \emph{Proceedings of the 61st Annual Meeting of the Association for Computational Linguistics (Volume 1: Long Papers)}, pages 14210--14228, Toronto, Canada. Association for Computational Linguistics.

\bibitem[{Shode et~al.(2022)Shode, Adelani, and Feldman}]{shode2022yosm}
Iyanuoluwa Shode, David~Ifeoluwa Adelani, and Anna Feldman. 2022.
\newblock yosm: A new yoruba sentiment corpus for movie reviews.
\newblock \emph{arXiv preprint arXiv:2204.09711}.

\bibitem[{Specia et~al.(2020)Specia, Blain, Fomicheva, Fonseca, Chaudhary, Guzm{\'a}n, and Martins}]{specia-etal-2020-findings-wmt}
Lucia Specia, Fr{\'e}d{\'e}ric Blain, Marina Fomicheva, Erick Fonseca, Vishrav Chaudhary, Francisco Guzm{\'a}n, and Andr{\'e} F.~T. Martins. 2020.
\newblock \href {https://aclanthology.org/2020.wmt-1.79} {Findings of the {WMT} 2020 shared task on quality estimation}.
\newblock In \emph{Proceedings of the Fifth Conference on Machine Translation}, pages 743--764, Online. Association for Computational Linguistics.

\bibitem[{Specia et~al.(2021)Specia, Blain, Fomicheva, Zerva, Li, Chaudhary, and Martins}]{specia2021findings}
Lucia Specia, Fr{\'e}d{\'e}ric Blain, Marina Fomicheva, Chrysoula Zerva, Zhenhao Li, Vishrav Chaudhary, and Andr{\'e}~FT Martins. 2021.
\newblock Findings of the wmt 2021 shared task on quality estimation.
\newblock In \emph{Proceedings of the Sixth Conference on Machine Translation}, pages 684--725.

\bibitem[{Wan et~al.(2022)Wan, Liu, Yang, Zhang, Chen, Wong, and Chao}]{wan2022unite}
Yu~Wan, Dayiheng Liu, Baosong Yang, Haibo Zhang, Boxing Chen, Derek~F Wong, and Lidia~S Chao. 2022.
\newblock Unite: Unified translation evaluation.
\newblock \emph{arXiv preprint arXiv:2204.13346}.

\bibitem[{Wang et~al.(2021{\natexlab{a}})Wang, Wang, Chen, Zhao, Luo, and Zhang}]{wang2021qemind}
Jiayi Wang, Ke~Wang, Boxing Chen, Yu~Zhao, Weihua Luo, and Yuqi Zhang. 2021{\natexlab{a}}.
\newblock Qemind: Alibaba's submission to the wmt21 quality estimation shared task.
\newblock \emph{arXiv preprint arXiv:2112.14890}.

\bibitem[{Wang et~al.(2021{\natexlab{b}})Wang, Shi, Wang, Zhang, Zhao, and Zheng}]{wang2021beyond}
Ke~Wang, Yangbin Shi, Jiayi Wang, Yuqi Zhang, Yu~Zhao, and Xiaolin Zheng. 2021{\natexlab{b}}.
\newblock Beyond glass-box features: Uncertainty quantification enhanced quality estimation for neural machine translation.
\newblock \emph{arXiv preprint arXiv:2109.07141}.

\bibitem[{Zerva et~al.(2022{\natexlab{a}})Zerva, Blain, Rei, Lertvittayakumjorn, C.~de Souza, Eger, Kanojia, Alves, Or{\u{a}}san, Fomicheva, Martins, and Specia}]{zerva-etal-2022-findings}
Chrysoula Zerva, Fr{\'e}d{\'e}ric Blain, Ricardo Rei, Piyawat Lertvittayakumjorn, Jos{\'e}~G. C.~de Souza, Steffen Eger, Diptesh Kanojia, Duarte Alves, Constantin Or{\u{a}}san, Marina Fomicheva, Andr{\'e} F.~T. Martins, and Lucia Specia. 2022{\natexlab{a}}.
\newblock \href {https://aclanthology.org/2022.wmt-1.3} {Findings of the {WMT} 2022 shared task on quality estimation}.
\newblock In \emph{Proceedings of the Seventh Conference on Machine Translation (WMT)}, pages 69--99, Abu Dhabi, United Arab Emirates (Hybrid). Association for Computational Linguistics.

\bibitem[{Zerva et~al.(2022{\natexlab{b}})Zerva, Blain, Rei, Lertvittayakumjorn, De~Souza, Eger, Kanojia, Alves, Ora{\v{s}}an, Fomicheva et~al.}]{zerva2022findings}
Chrysoula Zerva, Fr{\'e}d{\'e}ric Blain, Ricardo Rei, Piyawat Lertvittayakumjorn, Jos{\'e}~GC De~Souza, Steffen Eger, Diptesh Kanojia, Duarte Alves, Constantin Ora{\v{s}}an, Marina Fomicheva, et~al. 2022{\natexlab{b}}.
\newblock Findings of the wmt 2022 shared task on quality estimation.
\newblock In \emph{Proceedings of the Seventh Conference on Machine Translation (WMT)}, pages 69--99.

\bibitem[{Zhang et~al.(2020)Zhang, Kishore, Wu, Weinberger, and Artzi}]{Zhang_020BERTScore}
Tianyi Zhang, Varsha Kishore, Felix Wu, Kilian~Q. Weinberger, and Yoav Artzi. 2020.
\newblock \href {https://openreview.net/forum?id=SkeHuCVFDr} {Bertscore: Evaluating text generation with bert}.
\newblock In \emph{International Conference on Learning Representations}.

\end{thebibliography}
